\documentclass[10pt,journal,compsoc]{IEEEtran}

\ifCLASSOPTIONcompsoc
  \usepackage[nocompress]{cite}
\else
  \usepackage{cite}
\fi

\usepackage{xspace}
\usepackage[table,dvipsnames]{xcolor}

\usepackage{graphicx}
\usepackage{amsmath, mathtools}
\usepackage{amssymb}
\usepackage{booktabs}
\usepackage{multirow}
\usepackage{pifont}

\usepackage{siunitx}
\usepackage{setspace}
 \DeclareSIUnit\pixel{px}

\usepackage[format=plain,labelformat=simple,labelsep=period,font=small,compatibility=false]{caption}
\usepackage[font=footnotesize,skip=3pt,subrefformat=parens]{subcaption}

\usepackage{array}
\newcolumntype{L}[1]{>{\raggedright\let\newline\\\arraybackslash\hspace{0pt}}m{#1}}
\newcolumntype{C}[1]{>{\centering\let\newline\\\arraybackslash\hspace{0pt}}m{#1}}
\newcolumntype{R}[1]{>{\raggedleft\let\newline\\\arraybackslash\hspace{0pt}}m{#1}}

\usepackage[pagebackref=true,breaklinks=true,letterpaper=true,colorlinks,bookmarks=false]{hyperref}

\usepackage[capitalize]{cleveref}
\crefname{section}{Sec.}{Secs.}
\Crefname{section}{Section}{Sections}
\Crefname{table}{Table}{Tables}
\crefname{table}{Tab.}{Tabs.}
\usepackage{xr}

\makeatletter
\DeclareRobustCommand\onedot{\futurelet\@let@token\@onedot}
\def\@onedot{\ifx\@let@token.\else.\null\fi\xspace}

\def\eg{\emph{e.g}\onedot} 
\def\ie{\emph{i.e}\onedot}

\def\etal{\emph{et al}\onedot}
\makeatother

\newcommand{\reviewed}[1]{#1}
\renewcommand*{\and}{\hspace{0.9cm}}

\newcommand{\change}[1]{#1}

\usepackage{graphics}           
\usepackage{times}              
\usepackage{amsmath}            
\usepackage{amssymb}            
\usepackage{graphicx}
\usepackage{algorithm}
\usepackage[noend]{algpseudocode}
\usepackage{booktabs}
\usepackage{color}
\definecolor{instructioncolor}{rgb}{.5,.5,.5}

\usepackage[font=small]{caption}

\def\eqref#1{Eq.~(\ref{#1})}

\makeatletter
\usepackage{xspace}
\DeclareRobustCommand\onedot{\futurelet\@let@token\@onedot}
\def\@onedot{\ifx\@let@token.\else.\null\fi\xspace}
\def\eg{e.g\onedot} 
\def\ie{i.e\onedot}

\def\etal{{et al}\onedot}
\makeatother

\def\etalcite#1{\etal~\cite{#1}}

\usepackage{array}
\newcolumntype{L}[1]{>{\raggedright\let\newline\\\arraybackslash\hspace{0pt}}m{#1}}
\newcolumntype{C}[1]{>{\centering\let\newline\\\arraybackslash\hspace{0pt}}m{#1}}
\newcolumntype{R}[1]{>{\raggedleft\let\newline\\\arraybackslash\hspace{0pt}}m{#1}}

\renewcommand{\b}[1]{\mbox{\boldmath$#1$}}

\renewcommand{\v}[1]{{\b #1}} 

\begin{document}

\title{Towards Generating Realistic 3D Semantic Training Data for Autonomous Driving}

\author{
  Lucas Nunes \qquad
  Rodrigo Marcuzzi \qquad
  Jens Behley \qquad
  Cyrill Stachniss
  \IEEEcompsocitemizethanks{
    \IEEEcompsocthanksitem L. Nunes is with the Center for Robotics, University of Bonn, Germany, and the RWTH Aachen, Germany. E-Mail: lucas.nunes@igg.uni-bonn.de
    \IEEEcompsocthanksitem R. Marcuzzi, and J. Behley are with the Center for Robotics, University of
    Bonn, Germany. 
    E-mails: \{rodrigo.marcuzzi, jens.behley\}@igg.uni-bonn.de
    \IEEEcompsocthanksitem C. Stachniss is with the Center for Robotics, University of Bonn, Germany, and the Lamarr Institute for Machine Learning and Artificial Intelligence, Germany. E-Mail: cyrill.stachniss@igg.uni-bonn.de}
}

\IEEEtitleabstractindextext{%
  \begin{abstract}
    Semantic scene understanding is crucial for robotics and computer vision applications. In autonomous driving, 3D semantic segmentation
    plays an important role for enabling safe navigation. Despite significant advances in the field, the complexity of collecting and annotating 3D data is a bottleneck in this developments.
    To overcome that data annotation limitation, synthetic simulated data has been used to generate annotated data on demand.
    There is still, however, a domain gap between real and simulated data. More recently, diffusion
    models have been in the spotlight, enabling close-to-real data synthesis. Those generative models have been
    recently applied to the 3D data domain for generating scene-scale data with semantic annotations. Still, those methods either rely on image
    projection or decoupled models trained with different resolutions in a coarse-to-fine manner. Such intermediary representations impact the generated data quality due to errors added
    in those transformations. In this work, we propose a novel approach able to generate 3D semantic scene-scale data without relying
    on any projection or decoupled trained multi-resolution models, achieving more realistic semantic scene data generation compared to previous state-of-the-art
    methods. Besides improving 3D semantic scene-scale data synthesis, we thoroughly evaluate the use of the synthetic scene samples as labeled data to train a
    semantic segmentation network. In our experiments, we show that using the synthetic annotated data generated by our method as training data together with the real
    semantic segmentation labels, leads to an improvement in the semantic segmentation model performance. Our results show the potential of generated scene-scale
    point clouds to generate more training data to extend existing datasets, reducing the data annotation effort. Our code is available at \url{https://github.com/PRBonn/3DiSS}.\vspace{0.2cm}
  \end{abstract}
}

\maketitle

\IEEEdisplaynontitleabstractindextext
\IEEEpeerreviewmaketitle

\newlength{\teaserwidth}
\setlength{\teaserwidth}{0.24\textwidth}
\newlength{\teaserspace}
\setlength{\teaserspace}{0.05cm}

\section{Introduction} \label{sec:intro}


\IEEEPARstart{S}{emantic} perception is a key element in most robotics and computer vision applications. For example, in autonomous driving,
the capability of semantically interpret the scene is crucial for the safe navigation of the vehicle. Semantic segmentation is key
for scene understanding and has been intensively studied in recent years~\cite{milioto2019iros,zhu2021cvpr-caac, tang2020eccv}. Advances in this field
have been made possible also due to the availability of semantically annotated data~\cite{lin2014eccv,cordts2016cvpr}. However, the availability of such fine-grained annotated data
for 3D scene-scale data is  scarcer~\cite{behley2019iccv,geiger2012cvpr,caesar2020cvpr} compared to the image domain. Data collection and annotation of 3D data is more
challenging than for images, often hindering the scalability of 3D scene-scale labeled data.

\reviewed{Synthetic simulated data has been used to alleviate the labeling effort~\cite{ros2016cvpr,dosovitskiy2017corl,wilson2022ral}, but there is typically a domain gap between simulated and real data. This often impedes the large-scale use of simulated annotated data as labels for real-world applications~\cite{wu2019icra,chen2023iccv-ssaf,zhang2024iros-csud}. More recently, generative models have been in the spotlight given the recent advances in denoising diffusion probability models~(DDPMs)~\cite{ho2020neurips,dhariwal2021neurips,rombach2022cvpr,peebles2023iccv} and its realistic results. Due to the proximity between synthetic and real data that DDPMs enable, recent works have studied the use of synthetic data from those generative models as labeled data~\cite{tian2023neurips,sariyildiz2023cvpr,tian2024cvpr,geng2025arxiv}.
Still, those studies focus mainly on image data and on coarse downstream tasks, such as classification.}

\reviewed{Synthetic data generation has been studied for 3D data~\cite{luo2021cvpr,zeng2022neurips,xu2023cvpr-dzts} given the advances in the image domain~\cite{ho2020neurips,dhariwal2021neurips,rombach2022cvpr,peebles2023iccv}. For 3D scene-scale data the topic is also of great interest, even though the works in this field are more limited~\cite{lee2023arxiv,lee2024cvpr,ren2024cvpr,liu2024eccv}. } Some
works~\cite{lee2023arxiv,liu2024eccv} rely on a discrete formulation of the diffusion process, treating the scene as a fixed voxel grid,
and predicting the semantic label for each voxel. The memory cost of such methods is relatively high since empty voxels in the grid are also computed,
limiting the resolution of the generated scene. Other methods rely on intermediary data representations from the scene, such as image projections of the
3D point cloud~\cite{lee2024cvpr}, or training multiple models separately at different scene resolutions, modeling the generation process as a multi-resolution
conditional distribution~\cite{ren2024cvpr}. Despite these promising results, those methods rely on simpler or intermediate coarser representations of the real-world
data to simplify the 3D scene-scale data generation. Training the DDPM in such intermediate representations can limit its generation
capabilities due to the loss of information when creating those representations. In this work, we propose training the diffusion model without relying
on image projections or intermediate coarser scene generation. In our experiments, we show that
by training the DDPM directly on the 3D data and at the target resolution, we were able to achieve more realistic scene generation compared to previous state-of-the-art methods.


Alongside the proposed model, we extend the analysis from previous works in this field. We evaluate the use of the synthetic data as additional
labels for training a 3D semantic segmentation network, following the aforementioned works that used generated data as training data in the
image domain~\cite{tian2023neurips,sariyildiz2023cvpr,tian2024cvpr,geng2025arxiv}. We compare the use of synthetic scenes against real data as
training labels, and evaluate its use to extend the training set. We train the model with both real and synthetic scenes and assess the impact
of synthetic data in real-world applications. Besides, we compare the gaps between generated and real scenes, providing insights on the current
challenges in scene-scale 3D data generation.

\newpage
In sum, our main contributions are the following:
\begin{itemize}
  \item We propose a new method for generating 3D scene-scale semantic data without relying on multi-resolution representations or projections of the
  3D data.
  \item Our method generates more realistic scenes and this data is better suited for real-world tasks compared to previous state-of-the-art methods.
  \item We show that our generated data improves the performance on semantic segmentation when using both real and the synthetic data generated by our method.
  \item We compare both real and generated data distributions and identify current gaps between synthetic and real data to be addressed in future work.\vspace{-1em}
\end{itemize}

\section{Related Work} \label{sec:related_work}

\reviewed{\textbf{Denoising diffusion probabilistic models} (DDPMs) recently have been of great interest due to their high-quality image generation results~\cite{ho2020neurips,dhariwal2021neurips,rombach2022cvpr,zhou2023cvpr-sdft}. 
Such methods model the data generation as a denoising process, starting from Gaussian noise and iteratively predicting and removing noise from it,
slowly arriving at a novel sample from the target distribution.
Given the close-to-real generation of such models, recent studies focus on extending those methods to enable conditional generation, allowing the
synthesis of images resembling a specific input~\cite{ho2021neuripsws,zhang2023iccv,mo2024cvpr-ftsc}, often called conditioning. However, the DDPM
image generation is relatively slow compared to other generative methods,
\eg, Generative Adversarial Networks (GANs)~\cite{oeldorf2019icmla,karras2020cvpr,zhang2022cvpr-stgf} and Variational Auto-Encoders (VAEs)~\cite{vandenoord2017neurips,lee2022cvpr-aigu,huang2023cvpr-taic}. Therefore, recent works target reducing the DDPMs
inference time while maintaining the quality of the results, \eg, via distillation~\cite{salimans2022iclr,meng2023cvpr} or by analytically
solving multiple denoising steps~\cite{karras2022neurips,lu2022neurips}.}

\reviewed{\textbf{Diffusion models for 3D data} have been recently studied due to the advances in the image domain. Most of the methods in the field
focus on single object generation~\cite{zhou2021iccv-sgac,lyu2022iclr,zeng2022neurips,xu2023cvpr-dzts}. Those methods can be divided into two categories, point~\cite{zhou2021iccv-sgac,lyu2022iclr} and latent diffusion methods~\cite{zeng2022neurips,xu2023cvpr-dzts}.
The former operates directly over the point's 3D coordinates, learning how to generate the target point cloud from the starting random point coordinates. The latter
follows the latent diffusion model first proposed in the image domain~\cite{blattmann2022neurips,rombach2022cvpr}, where first, a VAE is trained to encode the point cloud into a latent space.
Then, the DDPM is trained to generate novel samples from the learned latent space, decoding it to a new sample.
More recent works extend such methods to operate in real-world LiDAR data. Some works~\cite{zyrianov2022eccv,nakashima2023arxiv,ran2024cvpr} use image projections from LiDAR scans
to train a DDPM to generate new samples of such image projections to then unproject it to the point cloud. Other works~\cite{nunes2024cvpr,zhang2024arxiv,song2024arxiv}
use a local point diffusion formulation~\cite{nunes2024cvpr} to enable scene-scale point diffusion, achieving LiDAR scene completion with DDPMs.}

\reviewed{\textbf{Semantic scene-scale diffusion models} target generating large-scale 3D scenes with semantic labels and have been studied in recent works~\cite{liu2024eccv,lee2024cvpr,ren2024cvpr,lee2023arxiv}. Lee~\etalcite{lee2024cvpr} tackle this problem by projecting the target semantic 3D data into a triplane image
representation. DDPM image models are then used to generate new triplane samples, which are then unprojected to the corresponding 3D scene. However, this data
projection loses information, limiting the details of the generated scene.
Ren~\etalcite{ren2024cvpr} achieve scene-scale data generation using a hierarchical approach
to model the coarse-to-fine nature of the scene. The generation process is formulated as a conditional distribution between different scene resolutions, training independently
a VAE and a DDPM for each pre-defined resolution. However, this formulation can lead to errors during the generation process since the finer
scene generation is unable to recover from mistakes done at coarser stages.
Distinctly, Lee~\etalcite{lee2023arxiv} propose using a discrete diffusion formulation to generate semantic scene-scale 3D data.
The scene is represented as a fixed 3D grid, and the DDPM is trained to generate the semantic labels for each cell in the grid, including an \textit{unoccupied} class.
Liu~\etalcite{liu2024eccv} extend this formulation by also leveraging a hierarchical multi-resolution approach, enabling a more detailed generation. Yet, such discrete
formulation limits the scene resolution due to memory constraints since the diffusion process is computed for the whole 3D grid, even though most of the cells are empty.}

\reviewed{In contrast to previous works, we propose a 3D latent diffusion model, which does not rely on data projections or decoupled multi-resolution models, but uses a single
sparse 3D VAE model to learn the target data distribution. By using a sparse 3D VAE, we train the model with the target data while avoiding the
increasing memory usage. Besides, we supervise the model to prune the scene within each decoder upsampling layers, learning the hierarchical scene structure with a single model, achieving more realistic generation.}

\reviewed{\textbf{Reducing labeling effort} has been the focus of many recent works given the challenge of scaling data annotation~\cite{dosovitskiy2017corl,wilson2022ral,he2020cvpr-mcfu,caron2021iccv}. To alleviate this problem, some works~\cite{ros2016cvpr,dosovitskiy2017corl,wilson2022ral} propose
using simulated training examples as labeled data. However, such simulated data come with a domain gap compared to the real-world data~\cite{wu2019icra,chen2023iccv-ssaf,zhang2024iros-csud}, often impeding the large-scale use of simulated training data.
Other works~\cite{he2020cvpr-mcfu,caron2021iccv,nunes2022ral,nunes2023cvpr} try to alleviate this problem using self-supervised pre-training methods to
learn a representation from unlabeled data, that allows to later fine-tuning the pre-trained model to the target downstream tasks. More recently, due to the realistic data synthesis
achieved by generative methods, some works~\cite{tian2023neurips,sariyildiz2023cvpr,tian2024cvpr,geng2025arxiv} have studied the use of generated samples as training data in the image domain.
Especially for the 3D data domain, the reduction of annotation requirements can largely impact the field due to the more complex task of data annotation compared to images. Previous
semantic scene-scale diffusion methods~\cite{lee2023arxiv,liu2024eccv,lee2024cvpr,ren2024cvpr} have motivated their works arguing about the capability of generating semantically annotated scenes on
demand. However, none of the previous works have studied the impact of using the
generated scenes as training data. Besides achieving more realistic scene generation, we evaluate the use of generated semantic scenes
as training data and show its potential to alleviate the burden of data annotation.}

In this work, we propose a semantic scene-scale diffusion method that can generate high-resolution scenes without intermediary
projections and using a single VAE model. We train a 3D sparse VAE to learn the target scene distribution while supervising it to prune unoccupied
voxels at each decoder upsampling layer. By doing so, the network learns to model the hierarchical nature of the scene with a single resolution VAE
while avoiding the increasing memory usage. Our experiments show that our method achieves more realistic scene
generation compared to previous state-of-the-art methods using a single sparse 3D~VAE. Additionally, we assess the use of our generated
scenes as training data for semantic segmentation. We show that by using our generated scenes together with real data, we improved
the semantic segmentation model performance. Finally, we perform an experiment to identify the gaps between real and generated data,
providing insights into the challenges to bridge this gap in future work.

\begin{figure*}
  \centering
  \includegraphics[width=0.9\linewidth]{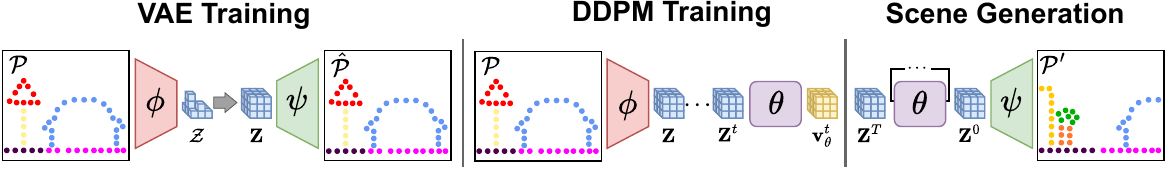}\vspace{-0.2cm}
  \caption{Our scene generation pipeline: First we train a VAE with the dense scenes $\mathcal{P}$ to
  reconstruct it as $\hat{\mathcal{P}}$ and to learn the sparse $\mathcal{Z}$ and dense $\v{Z}$ latent spaces. Next, the DDPM $\theta$ is trained over the latent $\v{Z}$
  sampling a random step $t$ to compute the noisy latent $\v{Z}^t$, training the model $\theta$ to predict $\textbf{v}_\theta^t$, following
  the \textbf{v}-parameterization formulation~\cite{salimans2022iclr}.
  Finally, novel scenes are generated by sampling random noise $\v{Z}^T \sim \mathcal{N}\left(\boldsymbol{0}, \boldsymbol{I}\right)$
  and denoising it with~$\theta$ over $T$ times, arriving to $\v{Z}^0 = \v{Z}_\theta$, decoding it with the VAE decoder to get to the generated scene~$\mathcal{P}' = \psi\left(\v{Z}_\theta\right)$.}\vspace{-0.4cm}
  \label{fig:pipeline_diagram}
\end{figure*}

\section{Our Approach} \label{sec:approach}

In this section, we describe our approach to generate semantic scene-scale data. We propose a DDPM~\cite{dhariwal2021neurips} model to generate semantic scene-scale data
without relying on intermediary image projections and without independent multi-resolution models training. We first train a VAE to encode the semantic
scene into a descriptive latent space. Then, we train a DDPM to learn the latent space of the VAE to generate new samples.
Using the VAE decoder to turn the sampled latent into the semantic scene. \cref{fig:pipeline_diagram} depicts an overview of our pipeline
for scene-scale generation.\vspace{-0.1cm}

\subsection{Semantic Scene VAE}

DDPMs are known for being computationally demanding. A common approach to simplify the generation process is to train an autoencoder
to learn a descriptive latent representation of the target data and then train the DDPM to generate novel samples from the learned latent space~\cite{rombach2022cvpr}.
Similarly, we first train a VAE with the semantic scene data, and then train the DDPM over the learned latent space, simplifying the semantic scene generation.

Let us define a voxelized 3D point cloud ${\mathcal{P}}$ with $N$ voxels with coordinates ${\mathcal{P}_\mathcal{C} = \{\v{c}_1,\dots, \v{c}_N\},\, \v{c}_n \in \mathbb{R}^3}$
within a fixed range over $[x,y,z]$, and features ${\mathcal{P}_\mathcal{F} = \{\v{f}_1,\dots, \v{f}_N\},\, \v{f}_n \in \mathbb{R}^4}$ as~${\v{f}_n = \left(x, y, z, s\right)}$
where $s \in \left\{1,\dots,C\right\}$ is the semantic~label for that voxel. We want to train a VAE with an encoder $\phi$ and a
decoder $\psi$, optimizing both such that~${\psi\left(\phi\left(\mathcal{P}\right)\right) = \hat{\mathcal{P}} \approx \mathcal{P}}$.

Ren~\etalcite{ren2024cvpr} modeled this training as $R$ independent sparse 3D VAE models trained at different resolutions of $\mathcal{P}$
as~${\hat{\mathcal{P}}_r = \psi_r\left(\phi_r\left(\mathcal{P}_r\right)\right)}$.
Then, $R$ independent DDPMs~$\theta_r$ are trained to hierarchically generate a novel scene
$\mathcal{P}'$ by conditioning the finer resolutions' scene generation to previous coarser stages
as~${\mathcal{P}'_r = \psi_r\left(\theta_{r-1}\left(\mathcal{N}\left(\v{0},\v{I}\right),\mathcal{P}'_{r-1}\right)\right)}$, until the final scene resolution.
The motivation behind this formulation is to model the coarse-to-fine scene data nature and to simplify the scene-scale data generation. However, we
argue that this coarse-to-fine data aspect can be modeled by a single encoder-decoder network in the sequential downsampling and upsampling layers. Furthermore,
this multi-resolution formulation can lead to incremental errors since the $R$~models are trained independently. Therefore, we train a single 3D
VAE model at the original resolution of $\mathcal{P}$.

The encoder $\phi$ receives the voxelized point cloud $\mathcal{P}$ as input and encodes it via consecutive downsampling convolutions into the latent representation
$\mathcal{Z}$ with coordinates~$\mathcal{Z}_\mathcal{C}=\{ \v{c}^\v{z}_1,\dots,\v{c}^\v{z}_M\}$ with $\v{c}_m^\v{z} \in \mathbb{R}^{3}$ and
features~${\mathcal{Z}_\mathcal{F}=\{ \v{f}^\v{z}_1,\dots,\v{f}^\v{z}_M\}}$ with~${\v{f}_m^\v{z} \in \mathbb{R}^{d_z}}$, where $M < N$, and $d_z$ being the latent feature dimension.
The latent coordinates~$\mathcal{Z}_\mathcal{C}$ are a lower resolution version of the input voxel coordinates~$\mathcal{P}_\mathcal{C}$ after the
encoder sparse convolutional layers. During the DDPM training, we want to generate novel scenes from scratch, not being constrained by a prior scene shape. Therefore,
we reshape the latent representation $\mathcal{Z}$ into a dense latent grid $\v{Z}$. Given that the voxelized point cloud $\mathcal{P}$ has a fixed range over the axes and a
fixed resolution of $0.1\,$m, the dense latent grid dimensions~$\v{Z} \in \mathbb{R}^{H \times W \times D \times d_z}$ are known. We initialize $\v{Z}$ with
zeros and set the occupied cells as $\v{Z}\left(\v{c}\right) = \mathcal{Z}\left(\v{c}\right) \,\forall\, \v{c} \in \v{Z}_\mathcal{C} \cap \mathcal{Z}_\mathcal{C}$.

\begin{figure}
  \centering
  \includegraphics[width=0.9\linewidth]{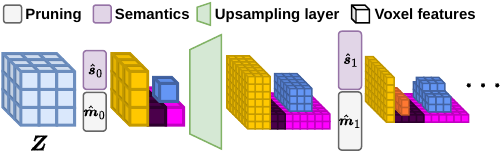}\vspace{-0.15cm}
  \caption{\reviewed{Diagram of the pruning process. Before each~$l^\textrm{th}$ upsampling layer the model predicts the semantics $\hat{\v{s}}_l$ and the pruning masks $\hat{\v{m}}_l$ and prune the unoccupied voxels, starting from the dense latent~$\v{Z}$.}}\vspace{-0.075cm}
  \label{fig:pruning_diagram}
\end{figure}

\reviewed{The goal of the decoder $\psi$ is to reconstruct $\mathcal{P}$ from the dense latent grid $\v{Z}$. To decode and upsample $\v{Z}$ would require many computational resources due to the exponential increase of the grid size within each upsampling. We overcome this issue by predicting a pruning mask $\hat{\v{m}}_l \in \left[0,1\right]^{H_l \times W_l \times D_l}$ and per-voxel semantic class $\hat{\v{s}}_l \in \mathbb{R}^{H_l \times W_l \times D_l \times C}$ with a pruning layer and a linear layer before~$l^\textrm{th}$ upsampling convolutional layer from $\psi$, as shown in \cref{fig:pruning_diagram}. That way, we upsample the latent grid $\v{Z}$ back to the original input resolution, pruning the unoccupied voxels while also predicting the semantics of each occupied voxel. At the same time, the supervision of the pruning masks $\hat{\v{m}}_l$ optimizes the network towards learning the scene layout at each~$l^\textrm{th}$ upsampling convolutional layer. We then compute $\v{Z} = \phi\left(\mathcal{P}\right)$, ${\hat{\mathcal{P}} = \psi\left(\v{Z}\right)}$, the predicted pruning masks $\hat{\mathcal{M}} = \{\hat{\v{m}}_1,\dots,\hat{\v{m}}_L\}$, and semantics $\hat{\mathcal{S}} = \{\hat{\v{s}}_1,\dots,\hat{\v{s}}_L\}$ from each~$l^\textrm{th}$ upsampling~layer.}\vspace{-0.1cm}

\subsubsection{Pruning Loss}
\reviewed{The pruning masks $\hat{\mathcal{M}}$ aim at predicting unoccupied voxels and remove them to reduce memory consumption when upsampling the dense latent $\v{Z}$, while learning the scene layout at each~$l^\textrm{th}$ upsampling layer. The goal is to learn the coarse-to-fine nature of the scene within a single
decoder, previously done with multiple independent VAE models~\cite{ren2024cvpr}. Therefore, we
supervise the prediction of the pruning masks~$\hat{\mathcal{M}}$ with the target masks $\mathcal{M}$, which are computed from the downsampling of the original scene $\mathcal{P}$.
We then train the model with the binary cross-entropy (BCE) and dice losses following recent mask-based segmentation approaches~\cite{cheng2021cvpr,marcuzzi2023ral,marcuzzi2023ral-meem}
computed between the prediction $\hat{\v{m}}_l$ and the target~$\v{m}_l$:}\vspace{-0.175cm}
\begin{equation}
  \label{eq:prune_loss}
  \mathcal{L}_{\text{prune}} = \sum_{l=1}^L \lambda_l \left(\mathcal{L}_{\text{bce}}^l\left(\hat{\v{m}}_l,\v{m}_l\right) + \mathcal{L}_{\text{dice}}^l\left(\hat{\v{m}}_l,\v{m}_l\right) \right),
\end{equation}
\noindent \reviewed{where $\lambda_l$ is the weight for pruning loss at the~$l^\textrm{th}$ upsampling layer. Both losses complement each other, while the BCE loss optimizes the
individual voxels learned features, the dice loss targets the scene layout, optimizing for the masks IoU. In that way, we optimize the decoder $\psi$ to learn to prune individual
voxels while also learning the whole scene layout at each~$l^\textrm{th}$ layer.}

\vspace{-0.2cm}\subsubsection{Semantic Loss}

\reviewed{We supervise the model to predict a semantic class for the voxels not pruned by the predicted pruning masks $\hat{\mathcal{M}}$. At each~$l^\textrm{th}$ upsampling layer, we compute the cross-entropy loss between the set of semantic predictions~$\hat{\v{s}}_l$ and the target semantics $\v{s}_l$ computed from the downsampling of the original scene input features~$\mathcal{P}_\mathcal{F}$. Given the class imbalance intrinsic to scene-scale data, first we supervise the VAE with the weighted cross-entropy~loss:\vspace{-0.15cm}
\begin{equation}
  \label{eq:sem_loss}
  \mathcal{L}_{\text{sem}} = - \sum^L_{l=1} \frac{1}{N^l} \sum^{N^l}_{n=1} w_{s_n^l} \log{\frac{\exp{\hat{s}^l_{n,s_n^l}}}{\sum^C_{c=1} \exp{\hat{s}^l_{n,c}}}},\vspace{-0.05cm}
\end{equation}
\noindent with $w_{s_n^l}$ as the weight of the target class at the $n^\textrm{th}$ voxel from the prediction $\hat{\v{s}}_l$. Then, for the second half of the training we train the model with the unweighted loss, \ie, $w_{s_n} = 1$. This unweighted loss training allows the VAE to learn the class imbalance from the training data to properly account for the scene data distribution.}

\vspace{-0.2cm}\subsubsection{Latent Loss}

With the pruning and semantic losses, we optimize the model to reconstruct structural and semantic information from the scene. However,
for the DDPM to generate meaningful scenes, the learned latent space has to be representative and continuous. Therefore, we regularize encoder
$\phi$ during training using the KL divergence to approximate the encoder latent distribution $q_{\phi}(\mathcal{P})$ to a target distribution, which is
set to $\mathcal{N}\left(\boldsymbol{0}, \boldsymbol{I}\right)$ as:
\begin{equation}
  \label{eq:latent_loss}
  \mathcal{L}_{\text{latent}} = \mathbb{KL}(q_\phi(\mathcal{P})\,||\,\mathcal{N}\left(\boldsymbol{0}, \boldsymbol{I}\right)),
\end{equation}
\noindent where $\mathcal{N}\left(\boldsymbol{0}, \boldsymbol{I}\right)$ is a zero-centered isotropic normal distribution.
With the VAE latent regularization we optimize the latent space towards a continuous representative distribution. This facilitates
the training of the DDPM and the generation of new samples from the latent distribution $q_\phi$, which later can be decoded by $\psi$.

\vspace{-0.2cm}\subsubsection{VAE training}

Finally, the VAE is trained to reconstruct the original scene while regularizing the learned latent space. The final
loss is a weighted combination of all three losses as:
\begin{equation}
  \label{eq:vae_loss}
  \mathcal{L}_\text{VAE} = \lambda_\text{prune}\mathcal{L}_\text{prune} + \lambda_\text{sem}\mathcal{L}_\text{sem} + \lambda_\text{latent}\mathcal{L}_\text{latent}.
\end{equation}

Besides, we do a refinement stage similar to previous approaches~\cite{ren2024cvpr,nunes2024cvpr}. After
the VAE is trained, we repeat the training adding noise to the latent $\v{Z}$, optimizing only the decoder~$\psi$ to reconstruct the scene $\mathcal{P}$
from the noisy latent. \reviewed{With \cref{eq:vae_loss}, by learning the pruning at each~$l^\textrm{th}$ layer, we achieved the coarse-to-fine modeling from the recent multi-resolution VAE approach~\cite{ren2024cvpr}
within a single model. In our experiments, we show that we~were able to generate more realistic scenes
compared to multi-resolution approaches, avoiding the problem from independent model training where the finer models inherit
mistakes from coarser stages.} \change{In addition to that, our method allows for more efficient scene generation, reducing inference time and memory requirements by modeling the problem with a single model. We report in the supplementary material experiments regarding inference time and memory requirements, comparing our method with hierarchical scene generation baselines.}

\vspace{-0.2cm}\subsection{Semantic Scene Latent Diffusion}

Given the training data distribution $p$ and the learned VAE latent distribution $q_\phi$, we want to train a diffusion model~$\theta$
to generate a novel dense latent from Gaussian noise as~${\v{Z}_\theta = \theta(\mathcal{N}\left(\boldsymbol{0}, \boldsymbol{I})\right)}$, such
that~${\v{Z}_\theta \sim q_\phi}$. Then, the VAE decoder~$\psi$ reconstructs the scene from the generated
latent~${\psi(\v{Z}_\theta)} = \mathcal{P}'$, where~$\mathcal{P}' \sim p$ is a novel scene reconstructed from the generated latent $\v{Z}_\theta$.

\subsubsection{Diffusion Training}

Denoising diffusion probabilistic models (DDPM) formulate data generation as a stochastic denoising process. A fixed number of~$T$ diffusion steps
is defined, and the training data is corrupted by iteratively sampling Gaussian noise and adding it to the data over the $T$ steps. Then,
the model is trained to predict the noise added at each one of the $T$ steps, iteratively removing the noise until arriving back to the uncorrupted
sample. During inference, Gaussian noise is sampled as the corrupted data at step $T$, and the model is used to denoise it, generating novel samples.

\textbf{Latent diffusion training} uses the trained VAE to encode a sample $\mathcal{P}$ from the target semantic scene data
distribution~$p$ into~$\v{Z} = \phi(\mathcal{P})$. Then, the DDPM training process operates directly on the VAE encoder dense
latent grid $\v{Z}$. At each training iteration, a random step $t$ is uniformly sampled from the $T$ diffusion steps.
Gaussian noise $\v{\epsilon}$ is sampled, and the corrupted data $\v{Z}^t$ at step $t$ is computed given pre-defined noise factors~${\beta_1,\dots,\beta_T}$,
with $\alpha_t = 1 - \beta_t$, and the cumulative product $\bar{\alpha}_t = \prod^t_{i=1} \alpha_i$ as:
\begin{equation}
  \v{Z}^t = \sqrt{\bar{\alpha}_t} \v{Z}^0 + \sqrt{1 - \bar{\alpha}_t}\v{\epsilon},\,\v{\epsilon} \sim \mathcal{N}(\v{0}, \v{I}).
\end{equation}

From the corrupted latent $\v{Z}^t$, the model $\theta$ is trained to predict the added noise $\v{\epsilon}$.
However, predicting only $\v{\epsilon}$ leads to slow convergence since the target distribution is only implicitly learned.
Salimans~\etalcite{salimans2022iclr} propose predicting~\textnormal{\textbf{v}} for faster convergence, defined in
terms of $\v{\epsilon}$, $\v{Z}^0$ and $t$. Thus, the model $\theta$ is optimized to learn the noise and target data distributions, with
\textnormal{\textbf{v}}$^t$ computed as:
\begin{equation}
  \label{eq:v_sample}
  \textnormal{\textbf{v}}^t = \sqrt{\bar{\alpha}_t}\v{\epsilon} - \sqrt{1 - \bar{\alpha}_t}\v{Z}^0,
\end{equation}
\noindent from which $\v{\epsilon}$ can be computed from $\textnormal{\textbf{v}}^t$ and $\v{Z}^t$ as:
\begin{equation}
  \label{eq:eps_from_v}
  \v{\epsilon} = \sqrt{\bar{\alpha}_t}\textnormal{\textbf{v}}^t + \sqrt{1 - \bar{\alpha}_t}\v{Z}^t,
\end{equation}
\noindent changing the model target without affecting the inference process. Therefore, the model $\theta$ is trained to predict $\textnormal{\textbf{v}}^t_\theta = \theta\left(\v{Z}^t, t\right)$:
\begin{equation}
  \label{eq:v_pred_loss}
  \mathcal{L}_\text{diff} = \lambda_{\text{SNR}\left(t\right)} \left\Vert \textnormal{\textbf{v}}^t - \textnormal{\textbf{v}}^t_\theta\right\Vert,\vspace{-0.25em}
\end{equation}
\noindent explicitly learning both the $\v{\epsilon}$ and $\v{Z}^0$ data distributions, where~$\lambda_{\text{SNR}\left(t\right)}$ is the
weight w.r.t. the signal-to-noise ratio (SNR) at the sampled diffusion step~$t$, following the Min-SNR-$\gamma$ weighting strategy~\cite{hang2023iccv}
to balance the training at the different diffusion steps.

\textbf{Latent diffusion inference} starts from random noise as the corrupted data as $\v{Z}^T \sim \mathcal{N}(\v{0},\v{I})$.
From $\v{Z}^T$ the model $\theta$ iteractively predicts $\textnormal{\textbf{v}}^t_\theta$ at each step $t$. From \eqref{eq:eps_from_v}, the predicted
noise $\v{\epsilon}^t_\theta$ added at step $t$ is computed from the prediction $\textnormal{\textbf{v}}^t_\theta$, iteratively removing
noise from $\v{Z}^T$ as:
\begin{equation}
  \v{Z}^{t-1} = \v{Z}^t - \frac{1 - \alpha_t}{\sqrt{1-\bar{\alpha}_t}}\v{\epsilon}^t_\theta + \frac{1 - \bar{\alpha}_{t-1}}{1 - \bar{\alpha}}\beta_t\mathcal{N}(\v{0},\v{I}),
\end{equation}
\noindent until arriving at the uncorrupted data sample $\v{Z}^0 = \v{Z}_\theta \sim q_\phi(\v{Z})$ from the VAE latent space. Then,
we decode the generated latent~$\v{Z}_\theta$ with the VAE decoder $\psi$, generating a novel semantic scene~$\mathcal{P}'$
from the target scene data distribution ${\psi(\v{Z}_\theta) = \mathcal{P}' \sim p}$.

\reviewed{\textbf{Conditioned latent diffusion} uses the diffusion model $\theta$ to generate samples conditioned to a target input data. In this context, we condition the
semantic scene generation to an input LiDAR scan to generate a dense and semantically annotated scene.
Following the classifier-free guidance~\cite{ho2021neuripsws}, we train the diffusion model $\theta$, conditioning it to a
LiDAR point cloud~${\mathcal{C} = \{\v{c}_1,\dots, \v{c}_P\}}$ with $\v{c}_p \in \mathbb{R}^3$ being the points coordinates. For the
conditioned training, the prediction $\textnormal{\textbf{v}}^t_\theta$ in \cref{eq:v_pred_loss} then becomes $\textnormal{\textbf{v}}^t_\theta = \theta(\v{Z}^t, \tilde{\mathcal{C}}, t)$,
with $\tilde{\mathcal{C}}$ having a probability $\rho$ of being the null token $\emptyset$ or the LiDAR point cloud $\mathcal{C}$
otherwise, \ie, switch between unconditioned and conditioned generation. We refer to the supplementary material for further details.}

\section{Experimental Evaluation} \label{sec:experiments}

\textbf{Datasets}. To generate our training data dense scenes, we use the SemanticKITTI dataset~\cite{behley2019iccv,geiger2012cvpr}, a 3D LiDAR semantic segmentation dataset
in the context of autonomous driving. The dataset provides sequences of 3D LiDAR scans with semantic labels. To generate the dense scenes,
we use an off-the-shelf SLAM~\cite{pan2024tro} method to compute scan poses throughout the whole sequence. With the labeled scans and the
corresponding poses, we aggregate the scan sequences with the labels, generating a voxel map of the scene with $0.1\,$m resolution.
We also use the labels to remove moving objects from the scene to avoid aggregating them and generating large artifacts from the moving objects
trajectories. 
 During training, we query a pose from one of the individual LiDAR scans and crop the corresponding dense scene from the map at the queried pose within a fixed range at $[x,y,z]$ between $[-25.6, -25.6, -2.2]$m and $[25.6, 25.6, 4.2]$m. For the semantic segmentation experiments, we use the same dataset, sequence $00$ from KITTI-360 dataset~\cite{liao2022pami}, and collected our own data with an Ouster LiDAR OS-1 with $128$ beams. We use LiDAR scans from KITTI-360 and our collected data as conditioning to generate the semantic dense scene from it, and to later use it as further labeled data to train a semantic segmentation model.

\textbf{Training}. As our VAE model, we use a 3D Sparse UNet~\cite{choy2019cvpr}. We train our VAE model for $50$ epochs, using the Adam optimizer~\cite{kingma2015iclr} with a learning rate of $10^{-4}$, multiplying it by $0.9$ every $5$ epochs. For the first $25$ epochs, we use the weighted cross-entropy loss with the class weights provided by the SemanticKITTI dataset~\cite{behley2019iccv,geiger2012cvpr}. For the last $25$ epochs, we use the unweighted cross-entropy. For the losses weights we use $\lambda_1=1.0$, $\lambda_2=1.0$, $\lambda_3=2.0$, $\lambda_4=3.0$ for the individual pruning layers losses in~\cref{eq:prune_loss}, with  $\lambda_\text{prune}=1.0$, $\lambda_\text{sem}=1.0$ and $\lambda_\text{latent}=0.002$ in~\cref{eq:vae_loss}. We trained the VAE with $6$ NVIDIA A40 GPUs. We use the same training scheme for the VAE refinement training.

For the DDPM, we use a 3D UNet model to learn the dense latent $\v{Z}$ and we train the model for $150$ epochs, using the AdamW optimizer~\cite{loshchilov2018iclr} with a learning rate of $2 \cdot 10^{-4}$, multiplying it by $0.8$ every $50$ epochs. The DDPM is trained with $\beta_T=0.015$ and $\beta_0=10^{-4}$, with $T=1000$, linearly interpolating it to get the noise factors $\beta_1, \dots, \beta_{T-1}$ in between. For the Min-SNR-$\gamma$ DDPM loss weighting, we use $\gamma=5$. We trained the DDPM with $8$ NVIDIA A40 GPUs. For the DDPM conditioned generation, we set the classifier-free guidance probability $\rho=0.1$ and conditioning weight to $2.0$. In the supplementary material, we provide further details regarding the VAE and DDPM models.

\begin{table}
  \centering
  \small
  \begin{tabular}{lcc}
    \toprule 
     & Eval. resolution [m] & MMD $\downarrow$ \\
    \midrule
    PDD~\cite{liu2024eccv} & 0.2 & 0.207 \\
    SemCity~\cite{lee2024cvpr} & 0.2 & 0.239 \\
    XCube~\cite{ren2024cvpr} & 0.2 & 0.160 \\
    Ours & 0.2 & \textbf{0.146} \\
    \midrule
    XCube~\cite{ren2024cvpr} & 0.1 & 0.101 \\
    Ours & 0.1 & \textbf{0.073} \\
    \bottomrule
  \end{tabular}
  \caption{Maximum mean discrepancy (MMD) between real and synthetic data.}
  \label{tab:mmd_eval}
\end{table}

\textbf{Baselines}. All baselines and our method are trained on the same set of dense scans. We compare our method with three baselines: SemCity~\cite{lee2024cvpr}, PDD~\cite{liu2024eccv} and XCube~\cite{ren2024cvpr}. All baselines are trained using their official implementations and default configurations. SemCity~\cite{lee2024cvpr} and PDD~\cite{liu2024eccv} are limited by the voxel resolution, with a maximum resolution of $0.2\,$m. For those baselines we downsample the training data to $0.2\,$m resolution. XCube~\cite{ren2024cvpr} and our method can generate scenes up to $0.1\,$m voxel resolution. Therefore, we compare all methods at $0.2\,$m resolution, downsampling XCube~\cite{ren2024cvpr} and our generated scenes, and compare our method with XCube~\cite{ren2024cvpr} at $0.1\,$m resolution.

\textbf{Semantic segmentation}. For the semantic segmentation training, we used the pipeline from our previous work~\cite{nunes2022ral} to train a 3D Sparse UNet. We train the model for $15$ epochs in all experiments with SGD optimizer with a learning rate of $0.24$ with a cosine annealing learning rate scheduler. For the experiments in \cref{sec:synth_quality}, we generate $8,000$ samples from each method and compute the metrics over those samples. For \cref{sec:sem_seg} and \cref{sec:real_plus_fake}, we generate the same amount of samples as SemanticKITTI dataset~\cite{behley2019iccv,geiger2012cvpr}, \ie, $19,130$ samples, to evaluate the performance with different subsets between real and generated samples. Also, since we remove the moving objects during the creation of the sequences maps, the occurrence of some classes decreases drastically, impacting the IoU over those classes. Therefore, in the mIoU computation, we ignore the classes that are mainly moving in the dataset, \ie, \textit{bicycle}, \textit{motorcycle}, \textit{person}, \textit{bicyclist} and \textit{motorcyclist}.

\subsection{Generated Scene Realism}
\label{sec:synth_quality}

\begin{table}
  \centering
  \small
  \begin{tabular}{lcc}
  \toprule 
     & Eval. resolution [m] & mIoU [\%] $\uparrow$ \\
    \midrule
    Validation set & 0.2 & 55.59 \\
    \midrule
    PDD~\cite{liu2024eccv} & 0.2 & 36.53 \\
    SemCity~\cite{lee2024cvpr} & 0.2 & 40.12 \\
    XCube~\cite{ren2024cvpr} & 0.2 & 40.02 \\
    Ours & 0.2 & \textbf{50.83} \\
    \midrule\midrule
    Validation set & 0.1 & 61.08 \\
    \midrule
    XCube~\cite{ren2024cvpr} & 0.1 & 27.24 \\
    Ours & 0.1 & \textbf{53.09} \\
    \bottomrule
  \end{tabular}
  \caption{mIoU over synthetic data with semantic segmentation model, trained on real data.}
  \label{tab:semseg_eval_over_synth}
\end{table}

\begin{figure*}
  \centering
  \includegraphics[width=0.95\linewidth]{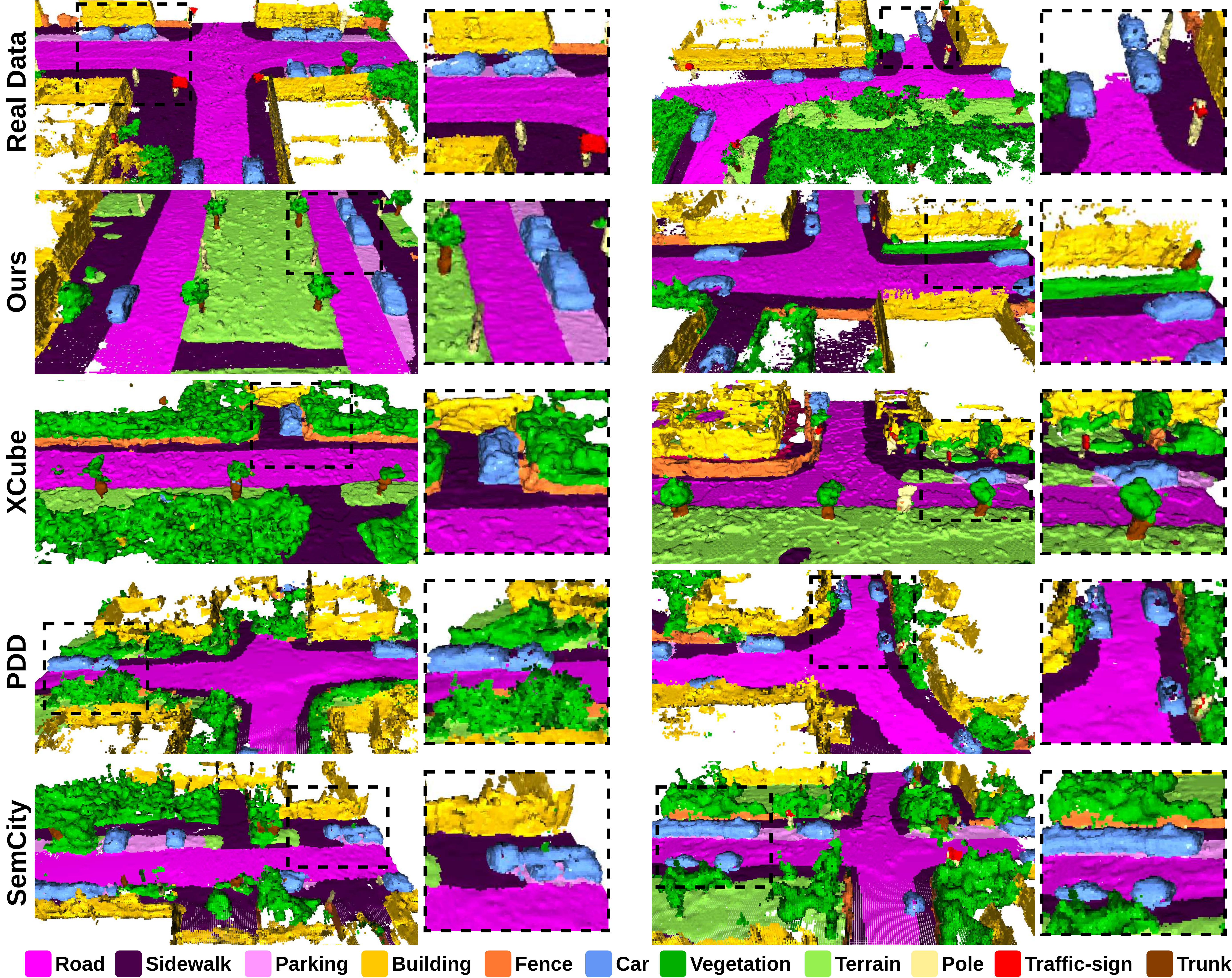}\vspace{-0.1cm}
  \caption{Comparison of real and unconditioned generated scenes from different methods. PDD and SemCity scenes are limited to $0.2\,$m resolution.
  The baselines scenes present rounder and too smooth shapes. Our method can generate more fine-grained details, closer to real data.}\vspace{-0.3cm}
  \label{fig:real_gen_comparison}
\end{figure*}

\begin{figure}
  \centering
  \vspace{-0.05cm}\includegraphics[width=0.9\linewidth]{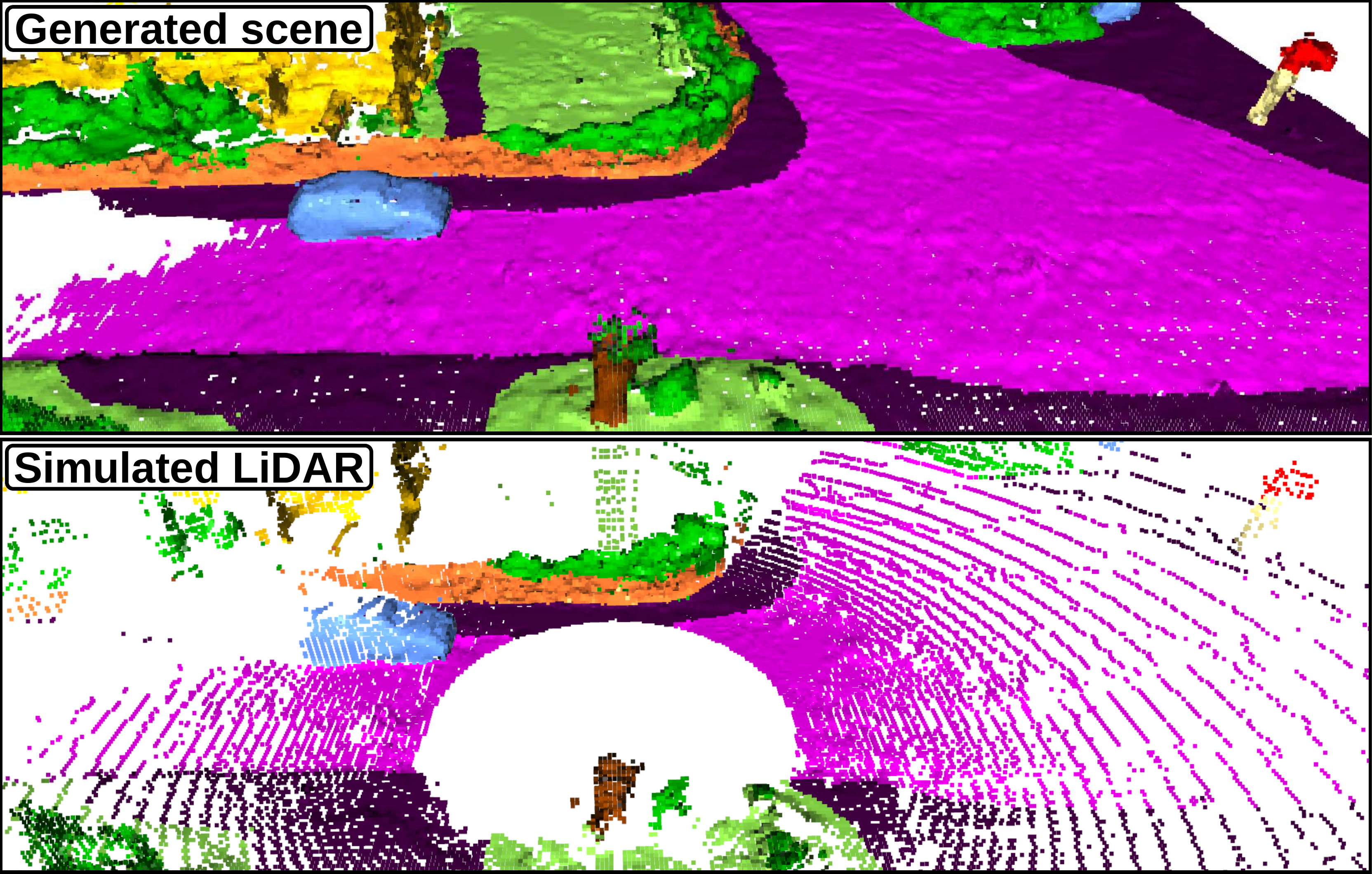}\vspace{-0.1cm}
  \caption{Simulated LiDAR point clouds from a dense generated scene.}
  \label{fig:lidar_sim}
\end{figure}

As the first experiment, we want to assess the quality of the generated scenes by evaluating how close they are to the real data. To do so, we first train a semantic segmentation model following the training protocol used in our previous work~\cite{nunes2022ral} with the real semantic scenes created through the scans aggregation from SemanticKITTI dataset~\cite{behley2019iccv,geiger2012cvpr}. Then, we input both real and synthetic scenes to the trained semantic segmentation network to compute their latent features, evaluating them with the maximum mean discrepancy (MMD). Besides, we follow the evaluation done by Liu~\etalcite{liu2024eccv}, using the trained semantic segmentation model to compute the mIoU over the generated scenes comparing it with the SemanticKITTI~\cite{behley2019iccv,geiger2012cvpr} validation set results. Given that the model was trained on real data, we expect higher mIoU over the generated scenes if they are similar to the real data in this evaluation. We do both evaluations at two resolutions, $0.1\,$m, and $0.2\,$m, since the PDD~\cite{liu2024eccv} and SemCity~\cite{lee2024cvpr} can only generate scenes up to $0.2\,$m~resolution.

\cref{tab:mmd_eval} shows the MMD metric between the generated and real scenes from SemanticKITTI dataset~\cite{behley2019iccv,geiger2012cvpr}. At $0.2\,$m resolution our approach and XCube~\cite{ren2024cvpr} surpass previous methods, showing the advantages of using latent diffusion and operating directly on the 3D data. At $0.1\,$m resolution, our method and XCube~\cite{ren2024cvpr} performance improves, still our method achieves the best performance. \cref{tab:semseg_eval_over_synth} depicts the results when computing the mIoU over the generated data with the semantic segmentation model trained on SemanticKITTI dataset~\cite{behley2019iccv,geiger2012cvpr}. At lower resolution the baselines' performance are similar while our method outperforms them, achieving a mIoU close to the real data performance. At higher resolution, XCube~\cite{ren2024cvpr} performance drops, while our method performance increases compared to the performance at $0.2\,$m resolution, showing that our method is able to generate scenes which are closer to the real data compared to the baselines.


\reviewed{\cref{fig:real_gen_comparison} shows scene examples from the real data compared with random scenes generated by our method and the baselines, where SemCity~\cite{lee2024cvpr} and PDD~\cite{liu2024eccv} generation is limited to $0.2\,$m resolution. In those examples, all baselines are able to learn the overall structure in the scene and generate reasonable scenarios. However, while the baselines generate smoother and rounder shapes, our approach is able to generated more fine-grained details closer to the real data. By avoiding intermediary representations and decoupled trained VAEs, our model can generate scenes with more details. These results show that our method generates closer-to-real scenes compared to the baselines. In the supplementary material we present additional experiments on the Waymo dataset~\cite{sun2020cvpr-sipf}, and also experiments regarding efficiency comparing our method with the hierarchical-models baseline XCube~\cite{ren2024cvpr}.}


\subsection{Generated Labels as Training Data}
\label{sec:sem_seg}

In this section, we aim at assessing the usability of the generated scenes as training data for downstream perception tasks.
We define subsets of training data varying the percentages of real data from SemanticKITTI dataset~\cite{behley2019iccv,geiger2012cvpr} available for training.
For each subset, we complement the real data with synthetic data such that the total amount of samples is the same as the full real data training set, \ie, $100$\% of data. Then, we
train the semantic segmentation network with both, the subsets of real data without any synthetic data, and the subsets complemented with generated data. We compare the results with
generated data from our method and XCube~\cite{ren2024cvpr}, the only two methods
able to generate scenes of $0.1\,$m resolution, computing the mIoU over the SemanticKITTI~\cite{behley2019iccv,geiger2012cvpr}
validation set. Besides evaluating it over the dense scenes, we also aim at evaluating the impact using sparse LiDAR data. Therefore, we simulate a LiDAR point cloud
from each dense scene from the real and generated scenes, as shown in \cref{fig:lidar_sim}, replicating the experiments also with the simulated LiDAR scans.

\begin{table}
  \centering
  \setlength{\tabcolsep}{0.15cm}
  \begin{tabular}{lccccc}
    \toprule
     & \multicolumn{5}{c}{mIoU [\%] $\uparrow$} \\
    \cmidrule{2-6}
     & \multicolumn{5}{c}{Real/Synth. [\%]} \\
     Synth. source & 10/90 & 25/75 & 50/50 & 90/10 & 100/0 \\
     \midrule
    Real only & 45.58 & 50.48 & 53.73 & 55.48 & \textbf{55.59} \\
    XCube~\cite{ren2024cvpr} & 49.28 & 52.12 & 55.04 & 55.75 & \textbf{55.59} \\
    Ours & \textbf{55.54} & \textbf{57.44} & \textbf{57.15} & \textbf{56.86} & \textbf{55.59} \\
    \bottomrule
  \end{tabular}
  \caption{Semantic segmentation performance for varying percentages of real data available to train the network, and complementing it with synthetic data
  to achieve 100\% size of the full training set trained with the LiDAR scans simulated from the densely generated scenes.}
  \label{tab:gt_synth_tradeoff_lidar}
\end{table}

\begin{figure}
  \centering
  \begin{subfigure}{.499\linewidth}
    \centering
    \includegraphics[width=1.\linewidth]{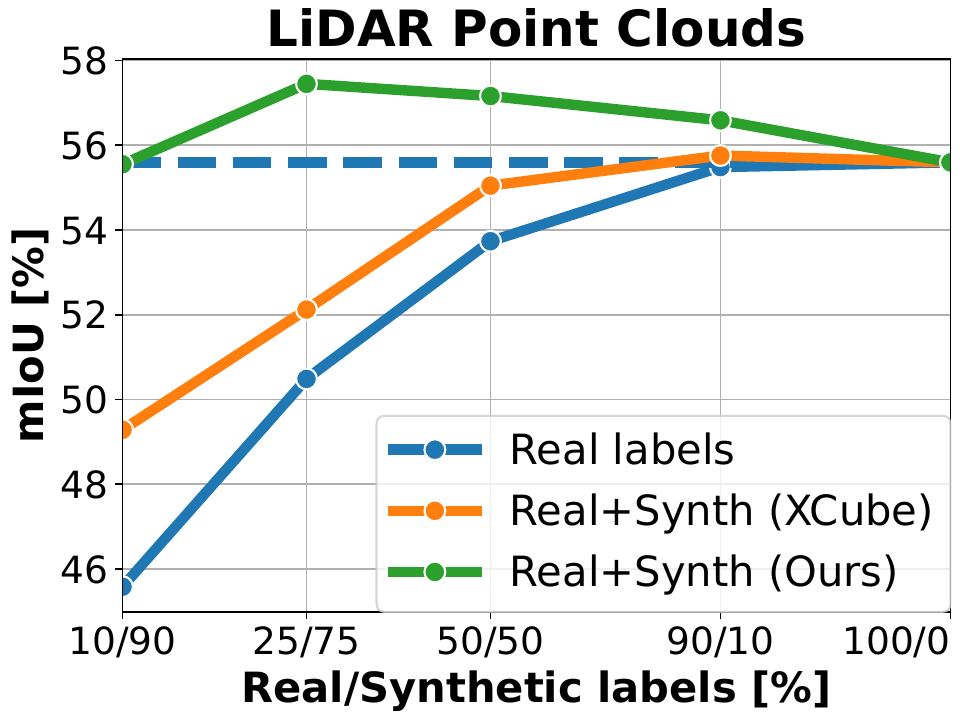}
    \caption{}
    \label{fig:gt_synth_lidar}
  \end{subfigure}\hspace{-0.12cm}
  \begin{subfigure}{.499\linewidth}
    \centering
    \includegraphics[width=1.\linewidth]{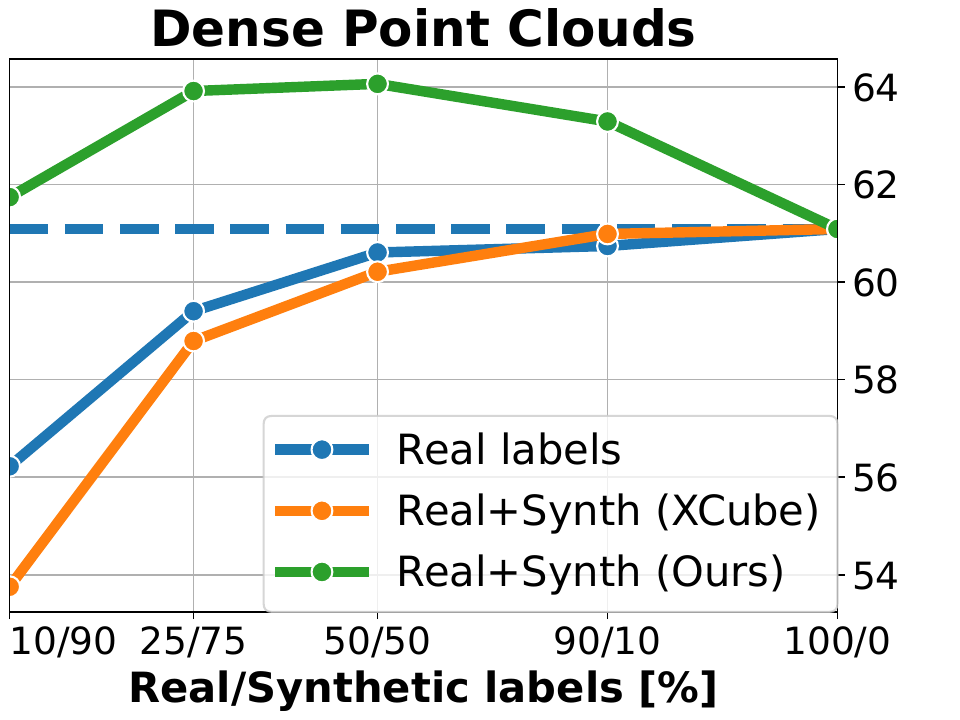}
    \caption{}
    \label{fig:gt_synth_dense}
  \end{subfigure}
  \caption{Semantic segmentation model performance trained with the different percentages of real data complemented with synthetic data
  from our model and XCube~\cite{ren2024cvpr}. (a) Model trained with LiDAR scans simulated from the dense scenes. (b) Model trained with dense scenes.}
  \label{fig:gt_synth}
  \end{figure}

\cref{tab:gt_synth_tradeoff_lidar} shows the results of the semantic segmentation network trained with the simulated LiDAR scans with varying percentages of real data complemented with generated data. As shown, the use of synthetic data improves the network performance whenever using scenes generated by XCube or by our method, converging to the same performance as the amount of real data gets closer to $100\%$. Still, the network trained with our synthetic data overall outperforms the network trained with the scenes generated by this baseline, as seen also in \cref{fig:gt_synth_lidar}. For dense point clouds in \cref{tab:gt_synth_tradeoff_dense}, the network trained with synthetic data generated by XCube achieves worse performance than the network trained only with real data. For such dense point clouds, the differences between the real and generated scene distributions have a higher impact on the network performance due to the amount of points in each generated sample. These results highlight the gap between the scenes generated by XCube and the real scenes. In contrast, when training with a subset of real data and with our generated scenes, the model achieved better performance than the network trained only with the full real training set. At first, this may sound counterintuitive due to the results presented previously in \cref{tab:semseg_eval_over_synth}, where our generated scenes presented a performance gap compared to the real data even though performing better than the baselines. Still, such improvement can be explained by the variability added to the training set when using generated data. The real scenes are collected sequentially, with few changes between consecutive point clouds, while the randomly generated scenes will be different from each other, adding variability to the training set. This increase in variability improves the network performance as long as enough real data is also seen by the model during training, as shown in \cref{fig:gt_synth_dense}. These results show that our method can generate scenes better suited to be used for downstream tasks compared to the baseline, and also show the potential of using generated samples as training data.

\change{Even though the experiments reported in the main paper are performed only on data from SemanticKITTI~\cite{behley2019iccv,geiger2012cvpr}, we argue that our approach is not limited to this dataset, as the annotated data generation is done over dense scenes, not limiting it to sensor- or dataset-specific characteristics. To support this argument, we refer to the supplementary material where we report experiments and ablations using the Waymo~dataset.}



\subsection{Synthetic Training Set Extension}
\label{sec:real_plus_fake}

In this section, we extend the experiment done in \cref{sec:sem_seg}. Given the real data training set, we would not expect to reduce the amount of real data available but rather enlarge
this dataset to achieve even better performance in a downstream task. Therefore, in this experiment, we use the scenes generated by our method to enlarge the
real data training set, evaluating the impact on the semantic segmentation model performance. We train the model with all the available
real training data while progressively adding synthetic data and evaluating the performance on the SemanticKITTI~\cite{behley2019iccv,geiger2012cvpr} validation set.
Given the results from previous sections, in this experiment we only evaluate the model using our generated data. As in \cref{sec:sem_seg}, we perform the experiment
twice, over the dense and the simulated LiDAR point~clouds.

\begin{table}
  \centering
  \setlength{\tabcolsep}{0.15cm}
  \begin{tabular}{lccccc}
    \toprule 
     & \multicolumn{5}{c}{mIoU [\%] $\uparrow$} \\
    \cmidrule{2-6}
     & \multicolumn{5}{c}{Real/Synth. [\%]} \\
     Synth. source &10/90 & 25/75 & 50/50 & 90/10 & 100/0 \\
    \midrule
    Real only & 56.22 & 59.40 & 60.60 & 60.73 & \textbf{61.08} \\
    XCube~\cite{ren2024cvpr} & 53.75 & 58.79 & 60.20 & 60.98 & \textbf{61.08} \\
    Ours & \textbf{61.73} & \textbf{63.91} & \textbf{64.06} & \textbf{63.28} & \textbf{61.08} \\
    \bottomrule
  \end{tabular}
  \caption{Semantic segmentation performance for varying percentages of real data available to train the network, and complementing it with synthetic data
  to achieve 100\% size of the full training set we train and evaluate with the densely generated point clouds.}
  \label{tab:gt_synth_tradeoff_dense}
\end{table}

\begin{table}
  \centering
  \begin{tabular}{ccc}
    \toprule 
    \multirow{2}{*}{\vspace{-0.2cm}\shortstack[c]{Additional\\synth. data [\%]}} & \multicolumn{2}{c}{mIoU [\%] $\uparrow$} \\
    \cmidrule(lr){2-3}
     & Dense & LiDAR \\
    \midrule
    0 & 61.08 & 55.59 \\
    10 & 62.42 & 55.96 \\
    25 & 62.47 & 56.70 \\
    50 & 62.97 & 56.88 \\
    75 & \textbf{64.14} & \textbf{57.77} \\
    100 & 64.07 & 56.53 \\
    \bottomrule
  \end{tabular}
  \caption{Semantic segmentation results when training the model with the full real data training set and adding additional synthetic data, both with
  dense point clouds and LiDAR scans simulated from the dense scenes.}
  \label{tab:gt_plus_synth_dense_lidar}
\end{table}

\begin{figure}
  \centering
  \begin{subfigure}{.499\linewidth}
    \centering
    \includegraphics[width=1.\linewidth]{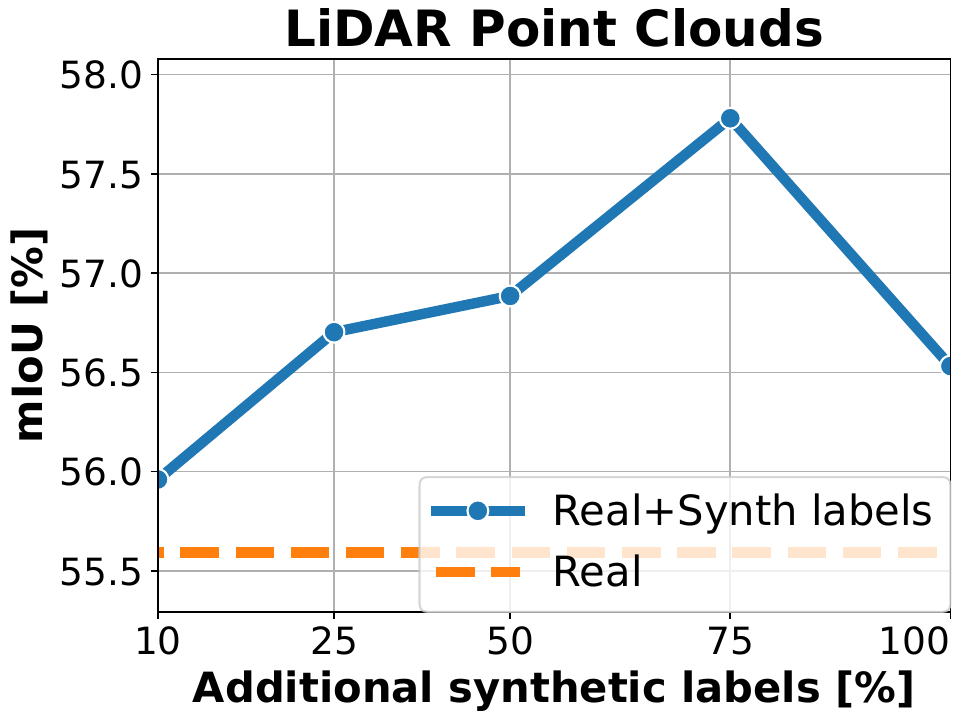}
    \caption{}
    \label{fig:gt_plus_synth_lidar}
  \end{subfigure}\hspace{-0.12cm}
  \begin{subfigure}{.499\linewidth}
    \centering
    \includegraphics[width=1.\linewidth]{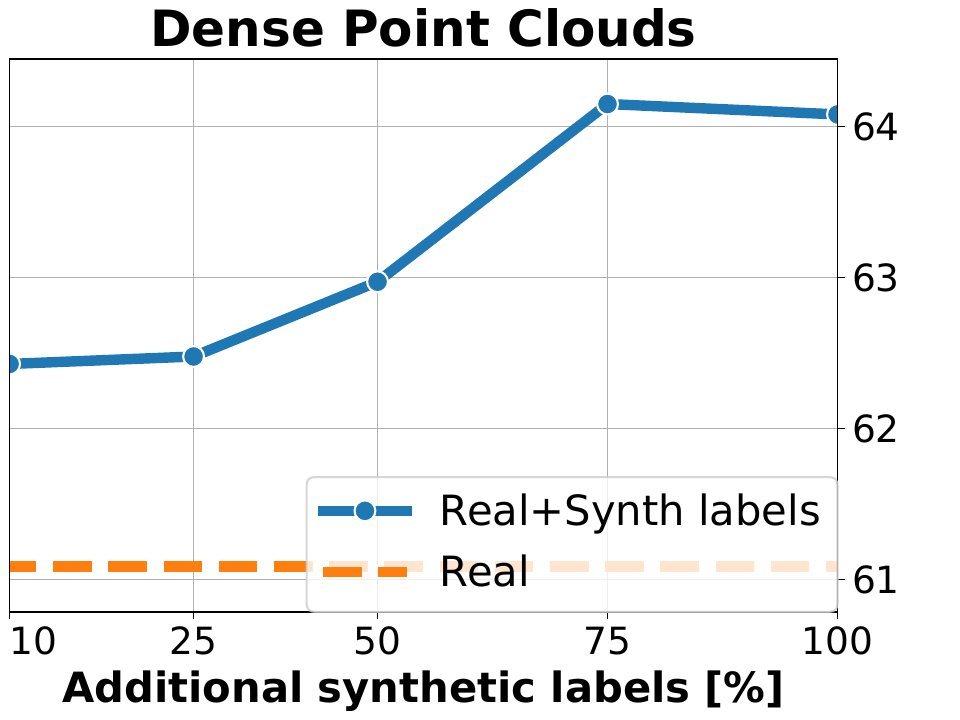}
    \caption{}
    \label{fig:gt_plus_synth_dense}
  \end{subfigure}
  \caption{Semantic segmentation model performance when trained with the full real training set and adding different amount of additional synthetic data
  from our model. (a) Model trained with LiDAR scans simulated from the dense scenes. (b) Model trained with dense scenes.}
  \label{fig:gt_plus_synth}
  \end{figure}

\cref{tab:gt_plus_synth_dense_lidar} shows the mIoU from the models trained with the full real training set and adding different percentages of the synthetic scenes
as additional data for both dense scenes and the LiDAR simulated point clouds. As shown, by using the synthetic scenes generated by our model, we were able to
improve the semantic segmentation model for both dense and LiDAR point clouds. As the amount of synthetic data increases, the model performance
also increases, saturating at $75\%$ additional synthetic labels, as seen in \cref{fig:gt_plus_synth}. When training with $100\%$ of the generated data, the model performance starts to degrade, although
still achieving better performance than the model trained only with real data. This can be explained by the fact that, even though better than previous
state-of-the-methods, our generated samples still have differences compared to the real data, as shown in \cref{tab:semseg_eval_over_synth}. Therefore,
by increasing the amount of synthetic data, the semantic segmentation model starts to learn more from the synthetic data distribution, and
consequently decreasing its performance on the real data.

\cref{tab:class_iou_lidar_real_plus_synth} and \cref{tab:class_iou_dense_real_plus_synth} show the class IoU when training only with the full real
training set compared adding $75$\% synthetic data to the real training set, which achieved the best performance. In
both cases, for LiDAR scans and for dense point clouds, the training with the synthetic data improves the performance in almost all the classes. These
results show the potential of synthetic data in helping to improve the performance in the target downstream task. Still, due to
a distribution gap between real and synthetic data discussed in \cref{sec:synth_quality}, there is space for improvement.

\subsection{Conditional DDPM as Data Annotator}
\label{sec:cond_gen}

\begin{figure*}
  \centering
  \includegraphics[width=0.95\linewidth]{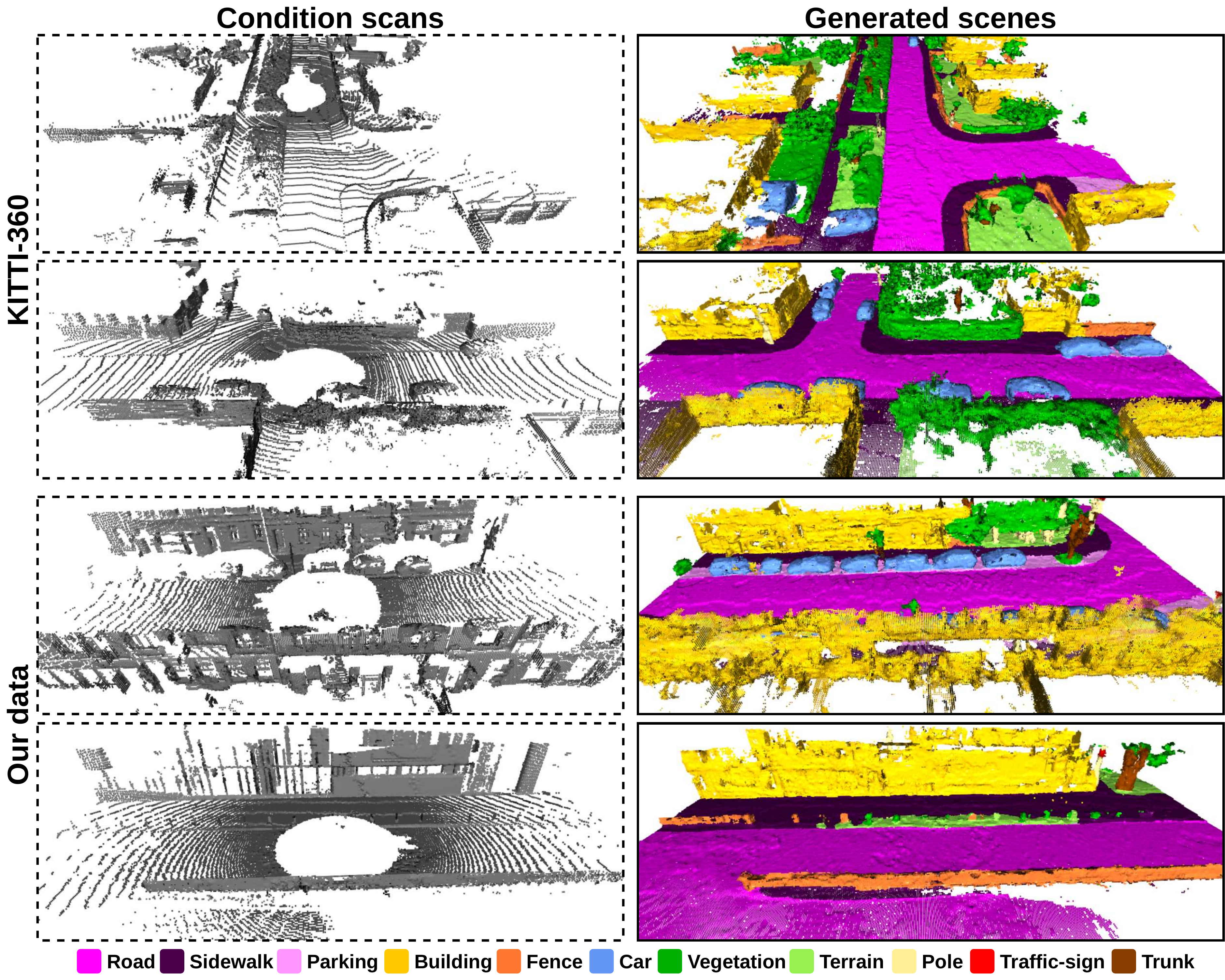}
  \caption{Generated scenes from our approach conditioned to KITTI-360~\cite{liao2022pami} scans and to our data collected with an Ouster LiDAR.}
  \label{fig:cond_samples}
\end{figure*}

Besides extending the training set with randomly generated samples as evaluated in \cref{sec:real_plus_fake}, one useful application of DDPMs is
to use conditional generation to annotate data. Target scenes could be
recorded with a LiDAR sensor and the annotated scene could be generated conditioned to the collected point clouds. In this case, the exhaustive manual data annotation
can be replaced with a simpler data curation, selecting only the most realistic scenes generated by the DDPM. Therefore, in this section, we train the DDPM with the
dense SemanticKITTI~\cite{behley2019iccv,geiger2012cvpr} scenes conditioned to their corresponding LiDAR scans from the dataset. With this conditional DDPM, we generate novel scenes conditioned
to sequence $00$ from KITTI-360~\cite{liao2022pami} and to our data collected with an Ouster LiDAR OS-$1$ with $128$ beams. Examples of conditional generated scenes are shown
in \cref{fig:cond_samples}. Next, we curate those conditional generated scenes by manually selecting scenes that look more realistic until arriving at
a total of $4,782$ point clouds, corresponding to approximately $25$\% of the size of the SemanticKITTI~\cite{behley2019iccv,geiger2012cvpr} training set. In
\cref{tab:number_of_scans_curation}, we provide information regarding the amount of scenes selected during the
curation process. Finally, we train the semantic segmentation~model with the SemanticKITTI~\cite{behley2019iccv,geiger2012cvpr} training set together with the curated
generated scenes to evaluate the use of conditioned DDPM as data annotator.

\begin{table*}
  \centering
  \small
  \begin{tabular}{lcccccccccccccc}
  \toprule
   & \multicolumn{13}{c}{IoU [\%] $\uparrow$} \\
  \cmidrule(lr){2-14}
  & car & truck & OV & road & park. & sidewalk & build. & fence & veg. & trunk & terrain & pole & sign \\
  \midrule
  Real only & 94.25 & 48.60 & 23.99 & 90.36 & 33.34 & 67.83 & 86.74 & 37.29 & 85.56 & \textbf{44.29} & 65.77 & \textbf{56.27} & \textbf{43.84} \\
  Real + 75\% Synth. & \textbf{94.67} & \textbf{59.81} & \textbf{28.45} & \textbf{91.28} & \textbf{42.02} & \textbf{70.14} & \textbf{87.59} & \textbf{42.36} & \textbf{86.14} & 40.57 & \textbf{67.38} & \textbf{56.27} & 42.17 \\
  \bottomrule
  \end{tabular}
  \caption{LiDAR simulated data class-wise IoU evaluated on the real data validation set comparing the network trained only with the full real
  training set with the training with the full training set with additional 75\% synthetic data generated with our method. OV refers to other-vehicle.}
  \label{tab:class_iou_lidar_real_plus_synth}
\end{table*}

\begin{table*}
  \centering
  \small
  \begin{tabular}{lcccccccccccccc}
  \toprule
   & \multicolumn{13}{c}{IoU [\%] $\uparrow$} \\
  \cmidrule(lr){2-14}
  & car & truck & OV & road & park. & sidewalk & build. & fence & veg. & trunk & terrain & pole & sign \\
  \midrule
  Real only & 93.77 & 42.24 & \textbf{49.00} & 92.16 & 46.52 & 69.24 & 89.28 & 41.08 & 88.07 & 51.88 & 65.81 & 61.12 & \textbf{58.36} \\
  Real + 75\% Synth. & \textbf{94.01} & \textbf{70.49} & 48.27 & \textbf{92.59} & \textbf{50.04} & \textbf{73.11} & \textbf{90.00} & \textbf{44.30} & \textbf{88.29} & \textbf{52.26} & \textbf{66.96} & \textbf{62.40} & 57.51 \\
  \bottomrule
  \end{tabular}
  \caption{Dense scenes class-wise IoU evaluated on the real data validation set comparing the network trained only with the full real
  training set with the training with the full training set with additional 75\% synthetic data generated with our method. OV refers to other-vehicle.}
  \label{tab:class_iou_dense_real_plus_synth}
\end{table*}

\begin{table}
  \centering
  \begin{tabular}{ccc}
    \toprule 
    Source & \#point clouds & \#curated scenes \\
    \midrule
    KITTI-360~\cite{liao2022pami} & 11,518 & 3,939 \\
    Our data & 5,276 & 843 \\
    \midrule
    Total & 16,794 & 4,782 \\
    \bottomrule
  \end{tabular}
  \caption{Number of point clouds in sequence $00$ from KITTI-360 and in our collected dataset and number of conditional generated scenes selected during the curation process.}
  \label{tab:number_of_scans_curation}
\end{table}

\begin{table}
  \centering
  \begin{tabular}{cccc}
    \toprule 
    \multicolumn{2}{c}{Data source} & \multicolumn{2}{c}{mIoU [\%] $\uparrow$} \\
    \cmidrule(lr){1-2}\cmidrule(lr){3-4}
    Real [\%] & Synth. [\%] & Dense & LiDAR \\
    \midrule
    100 & 0 & 61.08 & 55.59 \\
    100 & 25 & 62.47 & 56.70 \\
    100 & 75 & 64.14 & 57.77 \\
    \midrule
    100 & 25$^\dagger$ & \textbf{64.47} & \textbf{58.11} \\
    \bottomrule
  \end{tabular}
  \caption{Semantic segmentation trained with all the real SemanticKITTI training set together with unconditional randomly generated scenes and 
  with curated conditional generated scenes. $^\dagger$ refers to the curated conditional generated scenes.}
  \label{tab:gt_plus_curated}
\end{table}

\cref{tab:gt_plus_curated} shows the performance of the semantic segmentation network trained only with real data and with the $25$\% additional synthetic data
from unconditional and conditional curated generated scenes, also comparing with the best-performing model from \cref{tab:gt_plus_synth_dense_lidar} trained with additional
$75\%$ synthetic data. In this table, we notice that the network trained with the curated data achieves better performance than both uncurated training sets.
Even though using fewer samples, the model trained with only $25\%$ curated generated data achieves better performance than the model trained with additional $75$\% randomly
generated data. Those results show that the data curation process further improves the impact of the generated scenes on the network performance. This improvement
shows the potential of using conditioned DDPMs as data annotator, where the exhaustive manual data annotation could be replaced by a simpler data curation process.
Such an application could help to generate training data from specific scenarios, aiding the data annotation scalability.

\subsection{Generated and Real Data Gaps}
\label{sec:real_fake_gaps}

In previous sections, we performed experiments to investigate the potential of generated data to augment the real training set, improving the performance
of the models in the target downstream task. In this section, we aim at evaluating the opposite, comparing the data generated by our method with real data
to identify the gaps between both. For that, we first compute the class distributions from the real and generated data. Then, we use the semantic
segmentation network trained on real data to compute the IoU for each class in the generated and real scenes to identify the current gaps between synthetic and real
data.

\begin{figure}[t]
  \centering
  \includegraphics[width=0.9\linewidth]{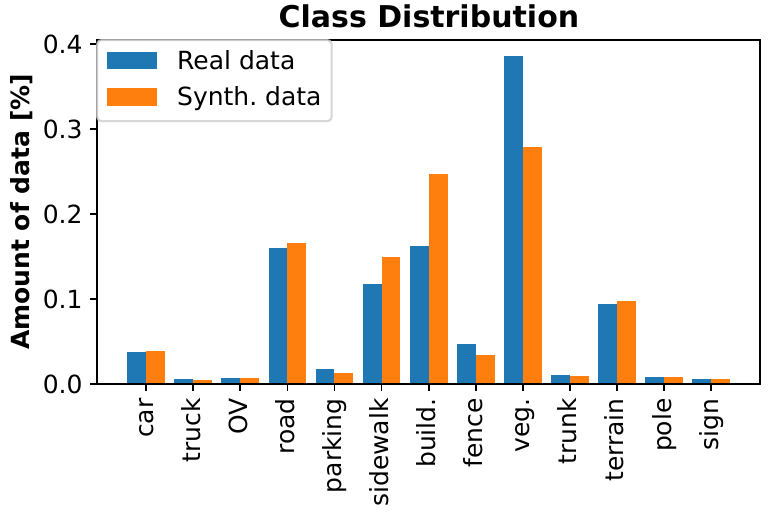}
  \caption{Class distribution over the real dataset compared with the synthetic data class distribution. (OV refers to other-vehicle class).}
  \label{fig:class_dist_hist}
\end{figure}

\begin{table*}
  \centering
  \small
  \begin{tabular}{lcccccccccccccc}
  \toprule
   & \multicolumn{13}{c}{IoU [\%] $\uparrow$} \\
  \cmidrule(lr){2-14}
  & car & truck & OV & road & park. & sidewalk & build. & fence & veg. & trunk & terrain & pole & sign \\
  \midrule
  Validation scenes & 93.77 & 42.24 & 49.00 & 92.16 & 46.52 & 69.24 & 89.28 & 41.08 & 88.07 & 51.88 & 65.81 & 61.12 & 58.36 \\
  Generated scenes & 92.29 & 10.42 & 24.10 & 92.48 & 33.74 & 77.34 & 89.75 & 47.73 & 81.81 & 42.99 & 61.72 & 43.55 & 28.79 \\
  \midrule
   Performance gap & \textcolor{red}{-1.48} & \textcolor{red}{-31.82} & \textcolor{red}{-24.90} & \textcolor{OliveGreen}{+0.32} & \textcolor{red}{-12.78} & \textcolor{OliveGreen}{+1.64} & \textcolor{OliveGreen}{+0.47} & \textcolor{OliveGreen}{+6.65} & \textcolor{red}{-6.26} & \textcolor{red}{-8.89} & \textcolor{red}{-8.89} & \textcolor{red}{-17.57} & \textcolor{red}{-29.57} \\
  \bottomrule
  \end{tabular}
  \caption{Class-wise IoU evaluated on real data validation set and synthetic data generated by our method with semantic segmentation model trained on real data, and the difference
  in performance between both real and synthetic data evaluation.}\vspace{-0.4cm}
  \label{tab:class_iou_real_synth}
\end{table*}

\cref{fig:class_dist_hist} shows the class distribution of SemanticKITTI~\cite{behley2019iccv,geiger2012cvpr} training set and of the scenes generated by
our method which was trained with SemanticKITTI~\cite{behley2019iccv,geiger2012cvpr}. As can be seen, the generated scenes class distribution follows the overall
distribution present in the dataset used to train the DDPM. Even though the generated scenes may increase the variability of the data compared to the real dataset,
as discussed in \cref{sec:sem_seg}, the amount of data per class remains close to the training data distribution. \cref{tab:class_iou_real_synth} shows the IoU
per class of the network trained with SemanticKITTI~\cite{behley2019iccv,geiger2012cvpr} training set and with the synthetic scenes
generated by our approach evaluated with the validation set, as well as the difference between both results. When comparing both \cref{fig:class_dist_hist} and \cref{tab:class_iou_real_synth}, we notice that the classes with high frequency in the synthetic scenes
in \cref{fig:class_dist_hist}, \eg, vegetation, road, sidewalk, fence, are also the classes that achieve
higher IoU in \cref{tab:class_iou_real_synth}. For those classes, the performance is comparable or even better compared to the evaluation on the real data validation set. At the same time, the less occurring
classes, \eg, truck, other-vehicle (OV), pole, traffic-sign, are the classes that perform worse in \cref{tab:class_iou_real_synth}, and for which there is a bigger performance gap
compared with the evaluation on the real validation set. The main gap between the generated and real data distributions therefore comes from those less occurring classes.
This analysis shows a direct relation between the amount of data samples per class and its generation quality. Hence, the DDPM
training would have to be balanced to account for the semantic classes imbalance present in scene-scale data to bridge this distribution gap in those
underrepresented classes. Even though we account for this class imbalance in the VAE training, during the DDPM training this is not
straightforward since the model is trained over the VAE latent, where there is no direct map between the semantic classes and the latent
features. Therefore, this class imbalance is not addressed in this article. Future works should address this class imbalance in the DDPM training, which would improve the generation quality of underrepresented classes,
increasing the generated scene quality.\vspace{-0.15em}

\section{Conclusion}
In this article, we propose a denoising diffusion probabilistic model able to achieve state-of-the-art semantic scene generation without relying on image projections or multi-resolution decoupled autoencoders trainings.
We follow recent latent diffusion methods and mask-based 3D segmentation approaches and train the model to learn the scene coarse-to-fine data nature within a single VAE
decoder upsampling layers. Our proposed method achieves closer-to-real scene generation compared to previous state-of-the-art methods
by avoiding intermediary representations and learning the scene data distribution at the original resolution. Besides, we performed
several experiments to assess the use of unconditional and conditional generated scenes as training data for real-world perception tasks.
In our experiments, we showed that by using the semantic scenes generated by our method together with the real data, we improve the model performance in
the semantic segmentation task. Our results show the potential of using DDPMs to enlarge the training set from unconditional random samples and
LiDAR conditioned generated scenes.
In our experiments, we also evaluated the relation between the semantic classes distribution between real and generated scenes and the generation
quality of each class. We show that the class imbalance inherent from the dataset directly impacts the class generation quality. From our analysis,
addressing this class imbalance during the DDPM training could help bridge the gap between real and generated scenes distributions. More realistic
scenes achieved by addressing this class imbalance can enable the large-scale use of generated scenes as training data, extending
the already available labeled data with random scenes and scenes conditioned to target scenarios, improving the performance on
real-world perception~tasks.

\change{\textbf{Limitations.} Despite the promising results, our training data generation is still affected by the removal of dynamic objects in the scene, leading to the removal of examples from potentially underrepresented classes, such as pedestrians, motorcyclists, and bicyclists. Besides, our evaluation mainly relies on the generation of annotated data from SemanticKITTI dataset. However, we expect that these findings are not limited to this dataset, but can also be extended to other datasets as suggested by the experiments on Waymo dataset reported in the supplementary material.} 


\appendices

\ifCLASSOPTIONcompsoc
  \section*{Acknowledgments}
\else
  \section*{Acknowledgment}
\fi
This work has partially been funded 
  by the Deutsche Forschungsgemeinschaft (DFG, German Research Foundation) under Germany's Excellence Strategy, EXC-2070 -- 390732324 -- PhenoRob,
  and by the German Federal Ministry of Research, Technology and Space~(BMFTR) under the Robotics Institute Germany~(RIG). The authors thank for the access to the Marvin cluster of the University of Bonn.

\bibliographystyle{IEEEtranSHack}
\bibliography{glorified, new}

\vspace{-10cm}\begin{IEEEbiography}[{\includegraphics[width=1in,height=1.25in,clip,keepaspectratio]{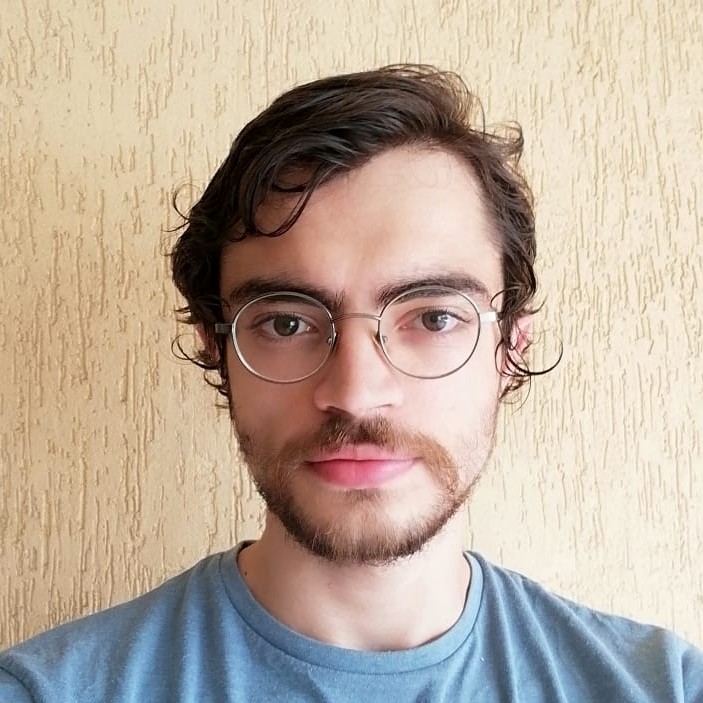}}]{Lucas Nunes} is a PhD student at the Photogrammetry and Robotics Lab at the University of Bonn since 2020. He obtained his MSc degree in Artificial Intelligence and Robotics from University of S\~ao Paulo in 2020, with a thesis on depth estimation of occluded regions. His area of interest lies in the area of perception for autonomous vehicles, representation learning for 3D LiDAR data, generative models for 3D scene-scale data. \vspace{-10cm}
\end{IEEEbiography}
\begin{IEEEbiography}[{\includegraphics[width=1in,height=1.25in,clip,keepaspectratio]{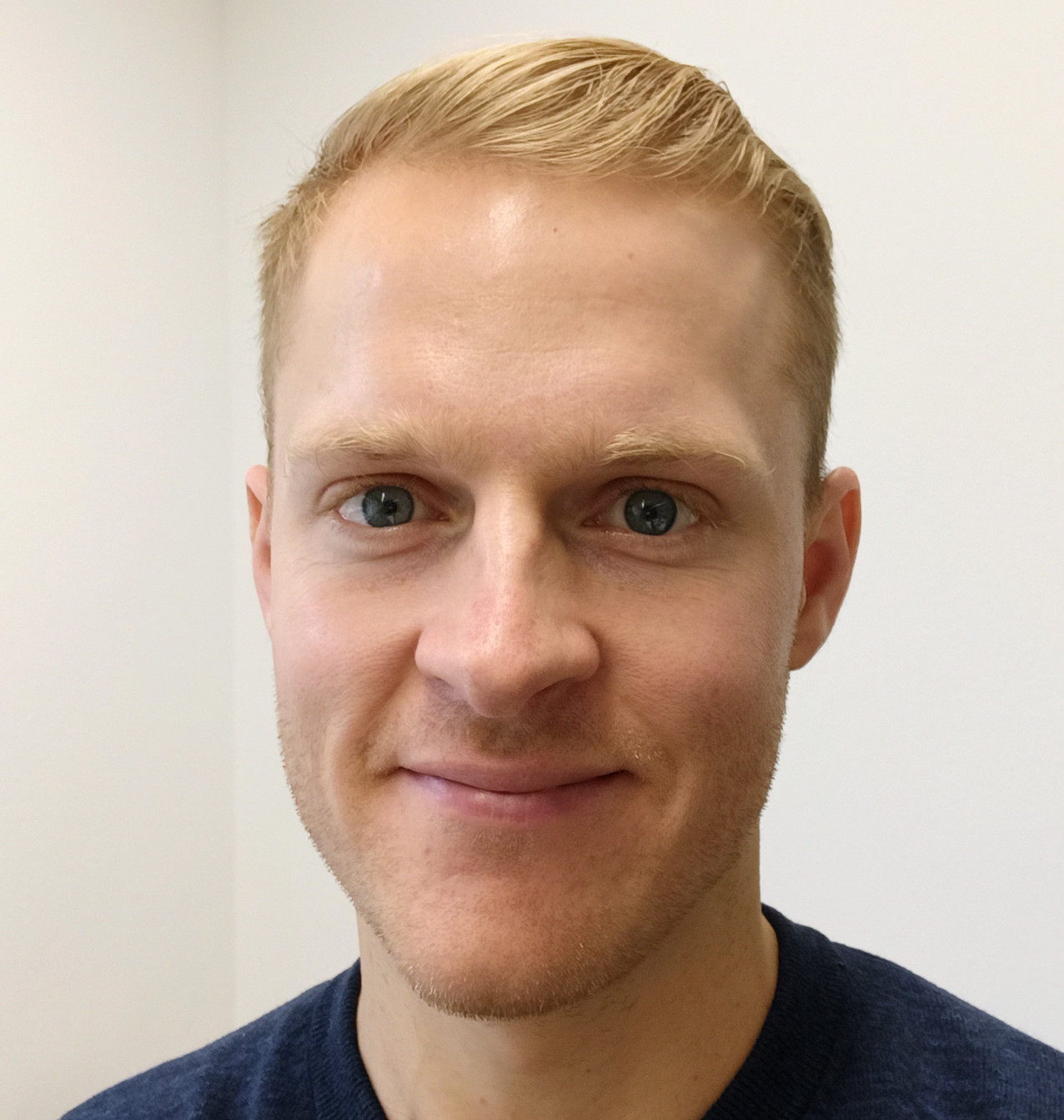}}]{Rodrigo Marcuzzi} is a PhD student at Photogrammetry and Robotics Lab at the University of Bonn since 2020. He obtained his Electrical Engineering Degree from National University of Rosario in 2019. During his studies he spent six months at IMT-Atlantique, France as an internship student. His area of interest lies in the area of 4D panoptic segmentation for autonomous vehicles, semantic occupancy prediction for autonomous vehicles. \vspace{-10cm}
\end{IEEEbiography}
\begin{IEEEbiography}[{\includegraphics[width=1in,height=1.25in,clip,keepaspectratio]{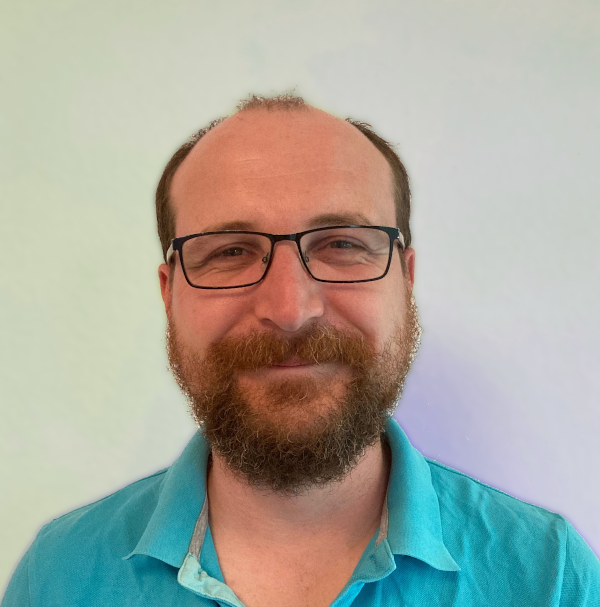}}]{Jens Behley} received his Ph.D. in computer science in  2014 from the Dept. of Computer Science at the University of Bonn, Germany. Since 2016, he is a postdoctoral researcher at the Photogrammetry \& Robotics Lab at the University of Bonn, Germany. He completed his habilitation at the University of Bonn in 2023. His area of interest lies in the area of perception for autonomous vehicles, deep learning for semantic interpretation, and LiDAR-based SLAM. \vspace{-10cm}
\end{IEEEbiography}
\begin{IEEEbiography}[{\includegraphics[width=1in,height=1.25in,clip,keepaspectratio]{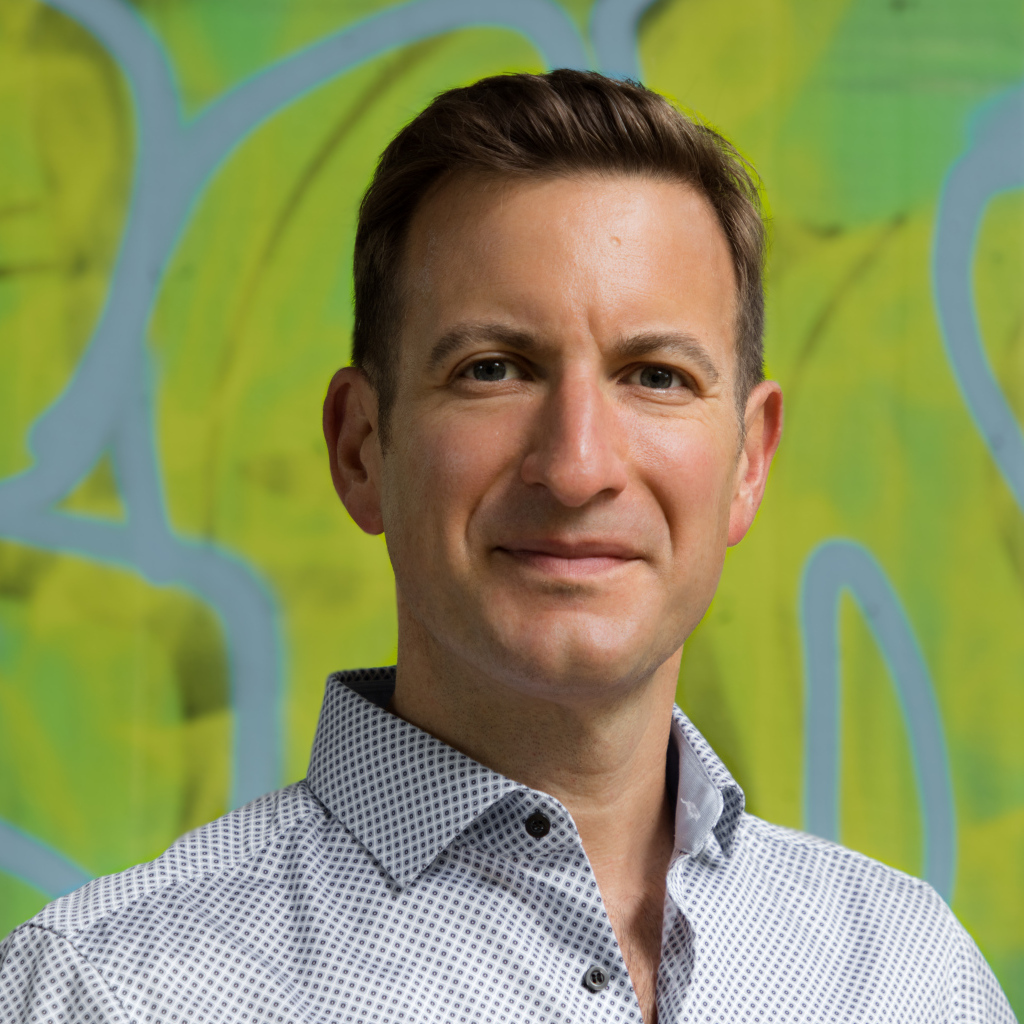}}]{Cyrill Stachniss} is a full professor at the University of Bonn, Germany, and with Lamarr Institute for Machine Learning and AI, Germany. He is the Spokesperson of the DFG Cluster of Excellence PhenoRob at the University of Bonn. His research focuses on probabilistic techniques and learning approaches for mobile robotics, perception, and navigation. Main application areas of his research are agricultural and service robotics and self-driving cars.
\end{IEEEbiography}

\end{document}


\title{\large\vrule depth 0pt height 0.5pt width 2.3cm\hspace{0.1cm}Supplementary Material\hspace{0.1cm}\vrule depth 0pt height 0.5pt width 2.3cm\\\Huge Towards Generating Realistic Autonomous Driving 3D Semantic Training Data}
\maketitle

\begin{figure*}
    \centering
    \includegraphics[width=0.95\linewidth]{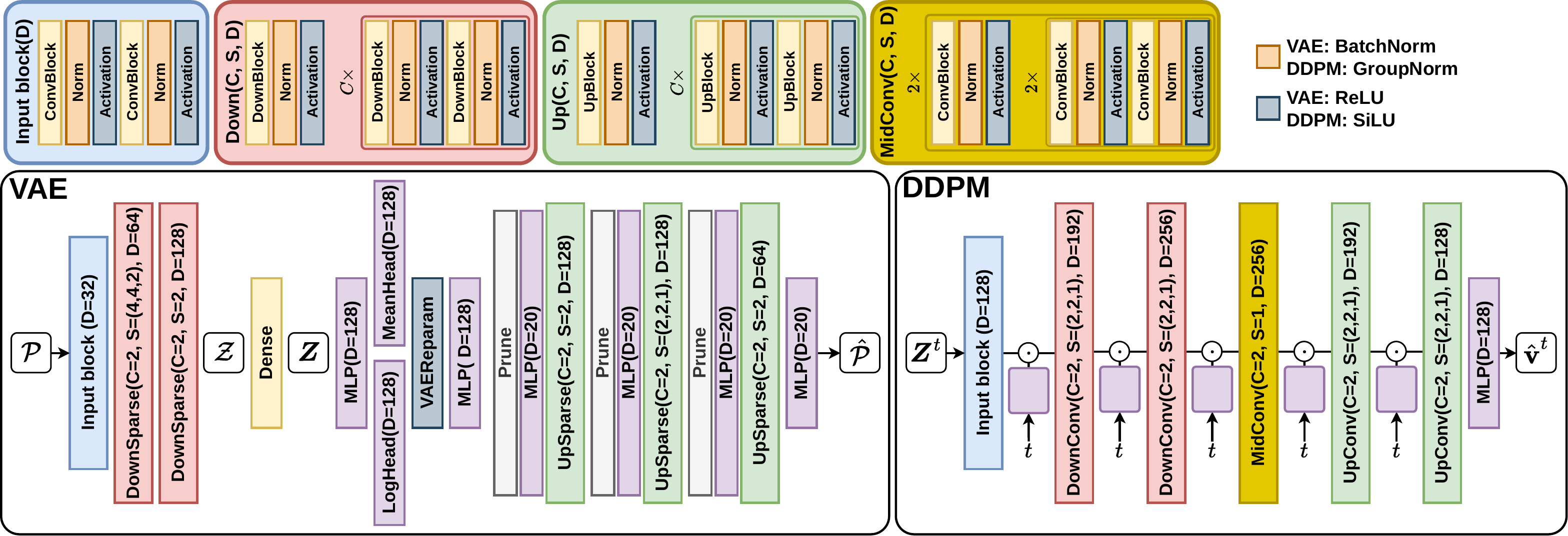}
    \caption{VAE and DDPM model architectures. The VAE receives the voxelized point cloud $\mathcal{P}$, encode it to the latent $\mathcal{Z}$ which
    is densified to $\v{Z}$ and decoded to $\hat{\mathcal{P}}$. The DDPM receives the noisy latent $\v{Z}^t$ at step $t$ and predicts $\hat{\textbf{v}}^t$
    following the \textbf{v}-parameterization formulation~\cite{salimans2022iclr}.}
    \label{fig:architecture}
\end{figure*}

\reviewed{This supplementary material provides further information regarding the method presented in the main article. \cref{sec:networks} provides further details about the network architectures used as the variational auto-encoder (VAE) and as the denoising diffusion probabilistic model (DDPM). \cref{sec:losses} provides further information regarding the binary cross-entropy (BCE) and dice losses used in the VAE training. \cref{sec:ablations} presents further assessments of our model regarding the pruning layers and efficiency compared to XCube~\cite{ren2024cvpr}, which employs the standard multi-model hierarchical scene generation strategy. \cref{sec:waymo} presents further experiments on Waymo dataset~\cite{sun2020cvpr-sipf}. \cref{sec:quantitative} shows the class IoU from the experiments done in Sec. 4.2 in the main article. \cref{sec:qualitative} presents additional qualitative results from conditional and unconditional generation from our method. Finally, \cref{sec:limitations} discuss current limitations of our proposed method.}

\section{Architectures}
\label{sec:networks}
This section presents the network architectures used in our approach. The VAE is trained to encode the scene $\mathcal{P}$ in the latent $\mathcal{Z}$, which is then padded to a dense grid $\v{Z}$ such that the VAE decoder can reconstruct $\hat{\mathcal{P}} \approx \mathcal{P}$ from it. The VAE architecture layers employ sparse operations to enable the processing of large-scale data and avoid exponential memory growth. The DDPM is trained with the VAE learned latent $\v{Z}$ by receiving as input the noisy latent $\v{Z}^t$ at step $t$, predicting $\textbf{v}^t$ following the \textbf{v}-parameterization formulation~\cite{salimans2022iclr}, where $\textbf{v}$ is parameterized in terms of the sampled noise $\v{\epsilon}$ and the uncorrupted data $\v{Z}^0$. Given the dense latent $\v{Z}$, the DDPM uses dense convolutional operations. Both model architectures are depicted in \cref{fig:architecture}.

\begin{figure*}
    \centering
    \includegraphics[width=0.9\linewidth]{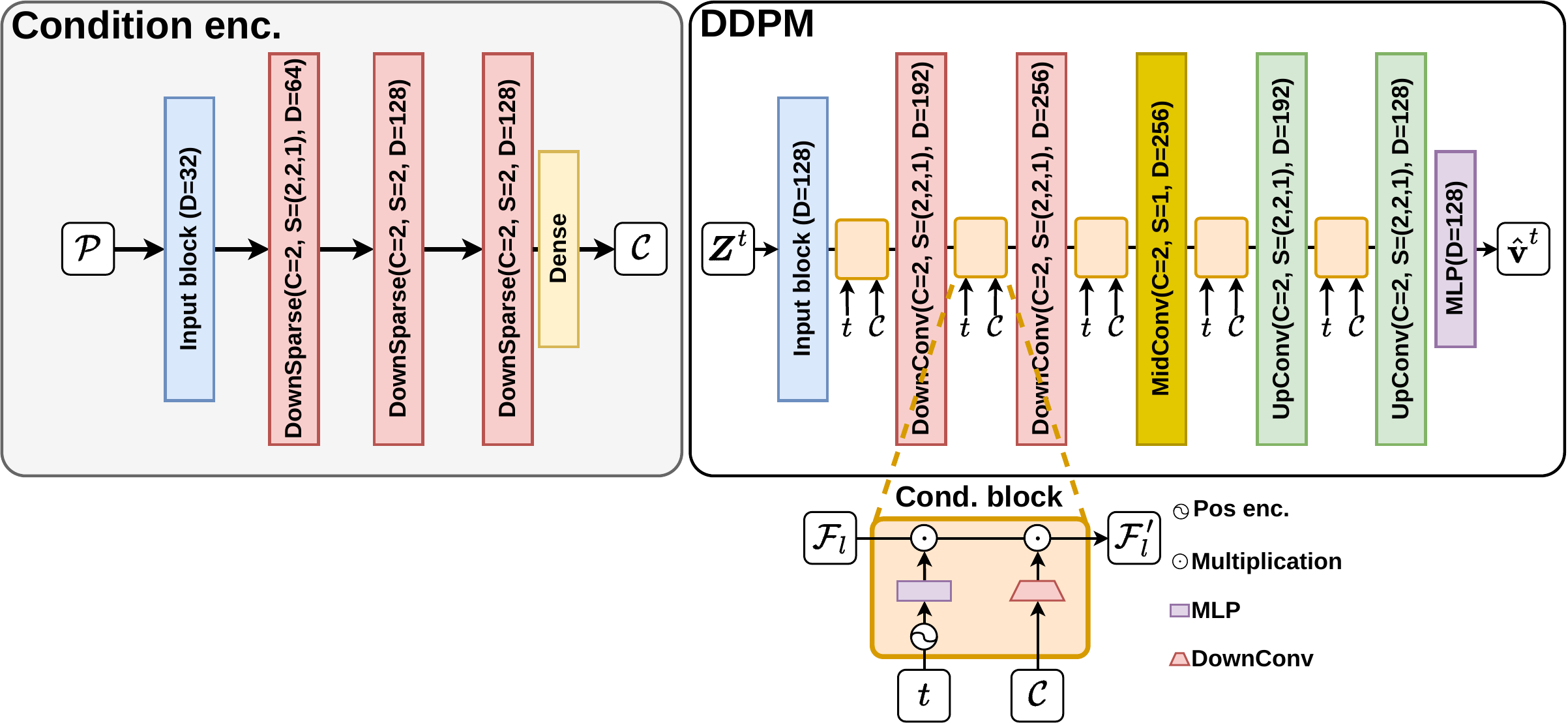}
    \caption{Condition encoder and DDPM model architecture. The condition encoder receives the voxelized point cloud $\mathcal{P}$, encode it to the condition latent token $\mathcal{C}$. The DDPM receives the noisy latent $\v{Z}^t$ at step $t$ and the condition latent token predicts $\mathcal{C}$. The condition block projects the step $t$ with an MLP and downsample the condition latent token $\mathcal{C}$ to the same latent resolution at the current DDPM layer, multiplying both into the layer feature $\mathcal{F}_l$ and computing the conditioned latent $\mathcal{F}_l'$ at each~$l^\textrm{th}$ layer in the DDPM.}\vspace{-0.2cm}
    \label{fig:architecture_cond}
\end{figure*}

\reviewed{The conditional DDPM model architecture is depicted in \figref{fig:architecture_cond}. The condition encoder computes a latent dense condition token $\mathcal{C}$ from the input point cloud $\mathcal{P}$. Then, at each~$l^\textrm{th}$ layer from the DDPM, the layer latent feature $\mathcal{F}_l$ is conditioned on the denoising step $t$ and the latent condition token $\mathcal{C}$. The step $t$ is first encoded by a positional encoder layer~\cite{nunes2024cvpr} and then projected to the current layer feature dimension with a linear layer, multiplying the projected feature by each element in the layer feature grid $\mathcal{F}_l$. The condition token $\mathcal{C}$ is projected to the current layer feature dimension and shape with a convolutional layer, merged to the layer latent grid $\mathcal{F}_l$ via an element-wise multiplication, and finally computing the conditioned feature grid $\mathcal{F}_l'$.}
    
\section{Pruning Losses}
\label{sec:losses}
This section details the $\mathcal{L}_\text{bce}$ and $\mathcal{L}_\text{dice}$ losses used to learn the pruning mask at each VAE upsampling layer.
As mentioned in Sec. 3.1 in the main article, the VAE is trained with a cross-entropy loss as the semantic loss and the KL divergence as the latent loss.
To learn the pruning masks, we follow recent mask-based segmentation approaches~\cite{carion2020eccv,cheng2021cvpr,marcuzzi2023ral,marcuzzi2023ral-meem},
supervising the mask prediction with the $\mathcal{L}_\text{bce}$ and $\mathcal{L}_\text{dice}$ losses. The $\mathcal{L}_\text{bce}$ aims at predicting
whether an individual voxel is occupied or not at a given~$l^\textrm{th}$ upsampling layer and is computed as:
\begin{equation}
    \label{eq:bce_loss}
    \mathcal{L}_{\text{bce}}^l = -{(\v{m}_l \log(\hat{\v{m}}_l) + (1 - \v{m}_l)\log(1 - \hat{\v{m}}_l))}.
\end{equation}
\noindent where $\v{m}_l$ is the target pruning mask and $\hat{\v{m}}_l$ is the predicted mask. At the same time, the dice loss aims at predicting the
whole scene layout by computing the dice coefficient, which approximates the IoU computation and maximizing this computation as:
\begin{align}
    \label{eq:dice_coeff}
    \text{dice}^l &= \frac{2 \left| \hat{\v{m}}_l \cap \v{m}_l \right|}{\left| \hat{\v{m}}_l \right| + \left| \v{m}_l \right|}, &\\
    \label{eq:dice_loss}
    \mathcal{L}_{\text{dice}}^l &= 1 - \text{dice}^l,
\end{align}
\noindent where as in \cref{eq:bce_loss}, $\v{m}_l$ is the target pruning mask and $\hat{\v{m}}_l$ the predicted mask.
Therefore, both losses complement each other, predicting both the individual voxel occupancy and the layout of the entire scene.\vspace{-0.5cm}

\section{Ablation Studies}
\label{sec:ablations}

\reviewed{In this section, we evaluate our proposed approach to support our choice of a convolutional DDPM model instead of standard transformer-based DDPM~\cite{peebles2023iccv}, and also evaluate our model in terms of efficiency. In \cref{sec:conv_vs_transformer}, we compare the quality of the scenes generated by our method with a convolution- and a transformer-based DDPM. \cref{sec:pruning_layers} compares the memory usage of our model with and without the pruning layers to show the advantage of learning and removing unoccupied voxels. Finally, \cref{sec:efficiency} evaluates the efficiency of modeling the coarse-to-fine scene generation within a single model in comparison to the standard hierarchical models strategy~\cite{bokhovkin2024arxiv,hua2025iccv,meng2025cvpr}.}

\subsection{Convolution vs Transformer Layers}
\label{sec:conv_vs_transformer}

\reviewed{In this experiment, we compare the quality of scenes generated by our method using a convolutional- and a transformer-based DDPM. \tabref{tab:trans_vs_conv} presents the MMD and mIoU results between both DDPM architectures trained with the same setup. As shown, the convolution-based DDPM outperformed the transformer DDPM in both metrics, especially in the mIoU experiment. At first, this seems counterintuitive given that recent generative approaches commonly rely on transformer-based architectures. However, this can be explained by the fact that for generating solely semantic labels, local information is more important than global context for generating locally consistent objects and structures. The underperformance of transformer-based DDPM can also be explained by the fact that such architectures require longer training and larger datasets. Nevertheless, both DDPM architectures outperformed the baselines, highlighting once more the advantages of our proposed method in comparison to prior works. Given those results, we decided for using a convolution-based DDPM in the experiments reported in the main paper.}

\begin{table}
  \centering
  \small
  \begin{tabular}{lcc}
    \toprule 
     & MMD $\downarrow$ & mIoU [\%] $\uparrow$ \\
    \midrule
    Transformer-based DDPM & 0.0806 & 40.83 \\
    Convolution-based DDPM & \textbf{0.0733} & \textbf{53.09} \\
    \bottomrule
  \end{tabular}
  \caption{\reviewed{MMD and mIoU between real and synthetic data comparing transformer- and convolution-based DDPMs.}}
  \label{tab:trans_vs_conv}
\end{table}

\subsection{Pruning Layers}
\label{sec:pruning_layers}

\reviewed{In this experiment, we aim to evaluate the gain in computational memory by using the pruning layer before each upsampling stage. \tabref{tab:prune_vs_noprune} presents the size of the latent features for each upsampling layer in our VAE model. Without pruning voxels, the memory requirement grows exponentially since all voxels, occupied and unoccupied, are still processed. When using the pruning layers, the model learns which voxels are unoccupied, removing them from the latent features, enabling a more efficient computation and avoiding an exponential increase in memory requirements. Without pruning, the large memory requirements make it intractable to process and generate scene at the target resolution of $0.1\,$m, requiring either reducing the resolution of the target scene or leveraging image projections as done by prior works~\cite{liu2024eccv,lee2024cvpr}.}

\begin{table*}
  \centering
  \small
  \begin{tabular}{lccccc}
    \toprule
     & \multicolumn{4}{c}{Feature size [MB]} & \multirow{2}{*}{\vspace{-0.2cm}\shortstack{Forward Pass\\Time [s]}} \\
     \cmidrule(lr){2-5}
     & Up Layer 1 & Up Layer 2 & Up Layer 3 & Up Layer 4 & \\
    \midrule
    Without pruning & 32.0 & 256.0 & 1056.0 & 4224.0 & 8.955 \\
    With pruning & 4.9 & 20.1 & 64.4 & 121.7 & 0.720 \\
    \bottomrule
  \end{tabular}\vspace{0.1cm}
  \caption{\reviewed{Per-layer comparison between the model with and without pruning, computing both feature sizes (MB) and forward pass time (seconds), averaged over 100 scene samples.}}
  \label{tab:prune_vs_noprune}
\end{table*}

\subsection{Computation Efficiency}
\label{sec:efficiency}

\reviewed{In this experiment, we assess the gain in efficiency of our coarse-to-fine scene generation modeling using a single model compared to standard multi-model hierarchical approaches~\cite{ren2024cvpr,bokhovkin2024arxiv,meng2025cvpr,hua2025iccv}. In this evaluation, we compute the inference time across $20$ samples using our proposed approach and the hierarchical baseline XCube~\cite{ren2024cvpr} and report the average inference time in \tabref{tab:inference_time}. As shown, in addition to achieving more realistic results (as reported in the main paper), our method is also more efficient in terms of computation time and memory requirements. As shown in \tabref{tab:inference_time}, our method requires about $3\times$ less time to generate a scene} \change{since it relies solely on a single stage to directly generate the high-resolution scene. In comparison, the hierarchical baseline performs the denoising diffusion process over each hierarchical stage, leading to slower inference.}

\change{In addition, our approach requires about $10\times$ fewer parameters compared to the hierarchical baseline. Besides the slower inference, the training of hierarchical decoupled models also leads to larger memory requirements given that multiple models are necessary to perform the scene generation. Our approach, in comparison, models the coarse-to-fine scene generation within a single model, allowing for a more efficient representation. This further highlights the advantage of our approach, achieving better performance while also being more efficient compared to the hierarchical scene generation approaches.}

\begin{table}
  \centering
  \small
  \vspace{0.1cm}\begin{tabular}{lccc}
    \toprule 
     & Inference [s] & \#params VAE & \#params DDPM \\
    \midrule
    XCube~\cite{ren2024cvpr} & 181.1 & 63.2M & 778.6M \\
    Ours & 58.3 & 6.5M & 66.2M \\
    \bottomrule
  \end{tabular}\vspace{0.1cm}
  \caption{\reviewed{Comparison of inference time and number of parameters between our method and the multi-model baseline XCube~\cite{ren2024cvpr}.}}
  \label{tab:inference_time}
\end{table}

\section{Waymo Dataset}
\reviewed{\label{sec:waymo}
To further assess the capabilities of our method, we train our model on the Waymo Open Dataset~\cite{sun2020cvpr-sipf}. We train the VAE for $14$ epochs and train the DDPM for $30$ epochs, using the same optimizer and learning rate as for the SemanticKITTI dataset~\cite{behley2019iccv,geiger2012cvpr}. Given that the Waymo dataset is about five times larger than the SemanticKITTI, we train the semantic segmentation network for $3$ epochs on Waymo to achieve roughly the same number of training iterations.}

\begin{figure*}[t]
  \centering
  \vspace{-0.05cm}\includegraphics[width=0.99\linewidth]{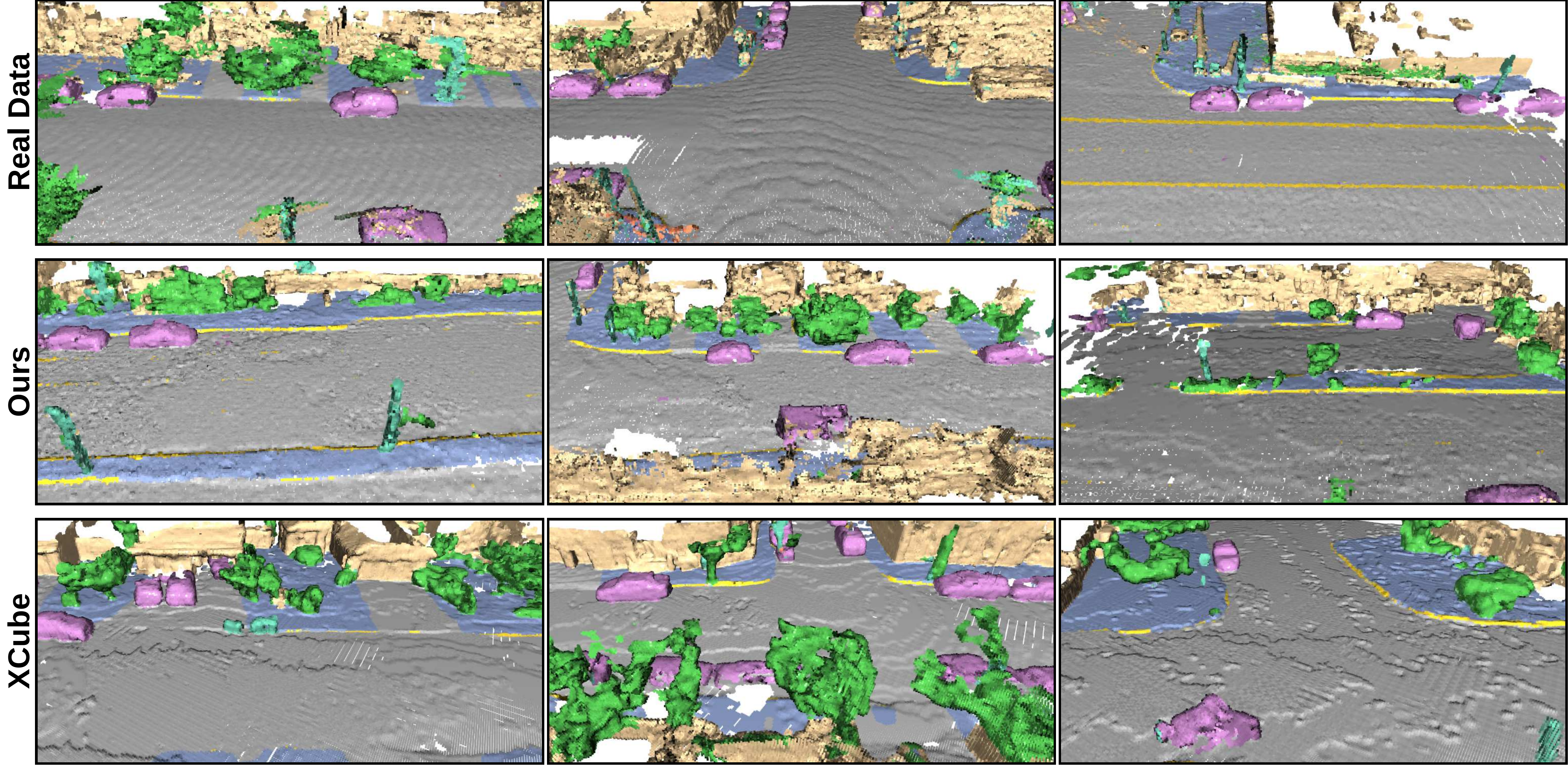}\vspace{-0.1cm}
  \caption{\reviewed{Qualitative comparison of Waymo results between groun truth and scenes generated by our method and XCube~\cite{ren2024cvpr}.}}
  \label{fig:waymo_qualitative}
\end{figure*}

\reviewed{We compare our method with XCube~\cite{ren2024cvpr} using the publicly available checkpoint for their model. We note that this comparison is not fair with our method, as our model was trained with limited GPU resources compared to the provided checkpoint. In the main paper, however, the assessment provided a fairer comparison given that all methods were trained with the same hardware specifications. Despite that, as shown in \cref{tab:mmd_eval} our method achieves competitive performance with the baseline even though it was trained with fewer GPU resources, and also while being more efficient as presented in \cref{tab:inference_time}. \cref{fig:waymo_qualitative} presents a qualitative comparison between the scenes generated by our method and XCube~\cite{ren2024cvpr}. As seen in the main experiments, the baseline generates smoother scenes, which might be visually appealing but are far from real scenes. Our method better imitates the sharper structures present in the real data, which makes it more suitable for use as training data, as shown in the main paper.} \change{This assessment with Waymo dataset \cite{sun2020cvpr-sipf} agrees with the experiments on SemanticKITTI present in the main paper. This suggests that the gain in performance when training the model with synthetic generated data is not limited to SemanticKITTI dataset, potentially also leading to improved performance in other datasets.} 

\begin{table}
  \centering
  \small
  \begin{tabular}{lcc}
    \toprule 
     & MMD $\downarrow$ & mIoU [\%] $\uparrow$ \\
    \midrule
    Validation set & - & 41.21 \\
    \midrule
    XCube~\cite{ren2024cvpr} & \textbf{0.033} & 39.10 \\
    Ours & 0.050 & \textbf{40.20} \\
    \bottomrule
  \end{tabular}\vspace{0.1cm}
  \caption{\reviewed{Maximum mean discrepancy (MMD) and mIoU evaluation between real and synthetic data on Waymo dataset.}}
  \label{tab:mmd_eval}
\end{table}

\vspace{0.3cm}\section{Quantitative Results}
\label{sec:quantitative}
In this section we provide further quantitative results regarding the class IoU from the scenes generated by our method and the baselines. In \cref{tab:class_iou_real_synth},
we report the class IoU from the model trained with real data evaluated on the real validation set and the scenes generated by the different methods.
Similar to Tab. 2 from the main article, we report the numbers at two resolutions, $0.2\,$m, and $0.1\,$m, since PDD~\cite{liu2024eccv} and SemCity~\cite{lee2024cvpr}
can only generate scenes up to $0.2\,$m resolution. Similar to what was presented in Tab. 2, at $0.2\,$m resolution, all the methods achieve similar performance
still our method achieves the best mIoU, as shown in Tab. 2. At $0.1\,$m resolution, our method surpasses the method able to generate scenes at this resolution,
achieving the best performance in all classes compared to the scene generated by the baseline. \cref{tab:class_iou_dense_real_plus_synth} depicts the class
IoU results from the experiment presented in Tab. 9 from the main article. As seen, with the curation process described in Sec. 4.4 the model performance achieves comparable to the best performing model, \ie, trained with additional $75\%$ of labels, even if using less synthetic training data. This suggests that the curation process can improve the model performance
when using synthetic scenes as training data by selecting only the most realistic scenes. This curation helps to bridge the gap between real and synthetic
data by removing less realistic samples.

\begin{table*}
\centering
\small
\begin{tabular}{lccccccccccccccc}
\toprule
    & \multicolumn{13}{c}{IoU [\%] $\uparrow$} \\
\cmidrule(lr){2-15}
    & Res. & car & truck & OV & road & park. & sidewalk & build. & fence & veg. & trunk & terrain & pole & sign \\
\midrule
Val. scenes & 0.2 & 91.10 & 45.90 & 31.73 & 87.76 & 34.87 & 58.18 & 86.33 & 25.32 & 86.64 & 42.79 & 64.11 & 54.75 & 47.72 \\
\midrule
PDD~\cite{liu2024eccv} & 0.2 & 80.24 & \textbf{11.81} & 0.29 & 81.14 & 8.99 & 57.00 & 79.19 & 34.82 & 78.71 & 7.96 & 54.65 & 15.67 & 0.81 \\
SemCity~\cite{lee2024cvpr} & 0.2 & 80.46 & 3.72 & 0.71 & 85.81 & 24.22 & 66.61 & 78.40 & 36.32 & \textbf{82.85} & 21.50 & 51.95 & 24.18 & 1.78 \\
XCube~\cite{ren2024cvpr} & 0.2 & 47.94 & 11.62 & 13.28 & 82.75 & 18.19 & 54.70 & 67.86 & 36.08 & 76.91 & 25.57 & 51.01 & 35.47 & 17.74 \\
Ours & 0.2 & \textbf{90.72} & 8.27 & \textbf{19.07} & \textbf{89.08} & \textbf{31.05} & \textbf{72.52} & \textbf{89.67} & \textbf{42.20} & 79.55 & \textbf{42.73} & \textbf{60.33} & \textbf{47.39} & \textbf{27.06} \\
\midrule
\midrule
Val. scenes & 0.1 & 93.77 & 42.24 & 49.00 & 92.16 & 46.52 & 69.24 & 89.28 & 41.08 & 88.07 & 51.88 & 65.81 & 61.12 & 58.36 \\
\midrule
XCube~\cite{ren2024cvpr} & 0.1 & 15.42 & 1.68 & 3.33 & 71.26 & 2.84 & 42.39 & 45.69 & 23.69 & 68.30 & 22.04 & 39.39 & 24.22 & 9.70 \\
Ours & 0.1 & \textbf{92.29} & \textbf{10.42} & \textbf{24.10} & \textbf{92.48} & \textbf{33.74} & \textbf{77.34} & \textbf{89.75} & \textbf{47.73} & \textbf{81.81} & \textbf{42.99} & \textbf{61.72} & \textbf{43.55} & \textbf{28.79} \\
\bottomrule
\end{tabular}
\caption{Class-wise IoU evaluated on real data validation set and synthetic data generated by the different methods with semantic segmentation model trained on real data.}
\label{tab:class_iou_real_synth}
\end{table*}

\begin{table*}
\centering
\small
\begin{tabular}{lcccccccccccccc}
\toprule
    & \multicolumn{13}{c}{IoU [\%] $\uparrow$} \\
\cmidrule(lr){2-14}
& car & truck & OV & road & park. & sidewalk & build. & fence & veg. & trunk & terrain & pole & sign \\
\midrule
Real only & 93.77 & 42.24 & 49.00 & 92.16 & 46.52 & 69.24 & 89.28 & 41.08 & 88.07 & 51.88 & 65.81 & 61.12 & \textbf{58.36} \\
Real + 75\% Synth. & 94.01 & 70.49 & 48.27 & \textbf{92.59} & \textbf{50.04} & \textbf{73.11} & \textbf{90.00} & \textbf{44.30} & \textbf{88.29} & 52.26 & \textbf{66.96} & \textbf{62.40} & 57.51 \\
Real + 25\% Synth.$^\dagger$ & \textbf{94.78} & \textbf{73.85} & \textbf{60.90} & 92.47 & 49.85 & 71.91 & 89.65 & 43.16 & 87.83 & \textbf{53.36} & 65.56 & 62.10 & 55.94 \\
\bottomrule
\end{tabular}
\caption{Dense scenes class-wise IoU evaluated on the real data validation set comparing the network trained only with the full real
training set with the training with the full training set with additional 75\% synthetic data generated with our method. OV refers to other-vehicle. $^\dagger$ refers to the curated conditional generated scenes.}
\label{tab:class_iou_dense_real_plus_synth}
\end{table*}

\section{Qualitative Results}
\label{sec:qualitative}
\begin{figure*}[t]
    \centering
    \includegraphics[width=0.99\linewidth]{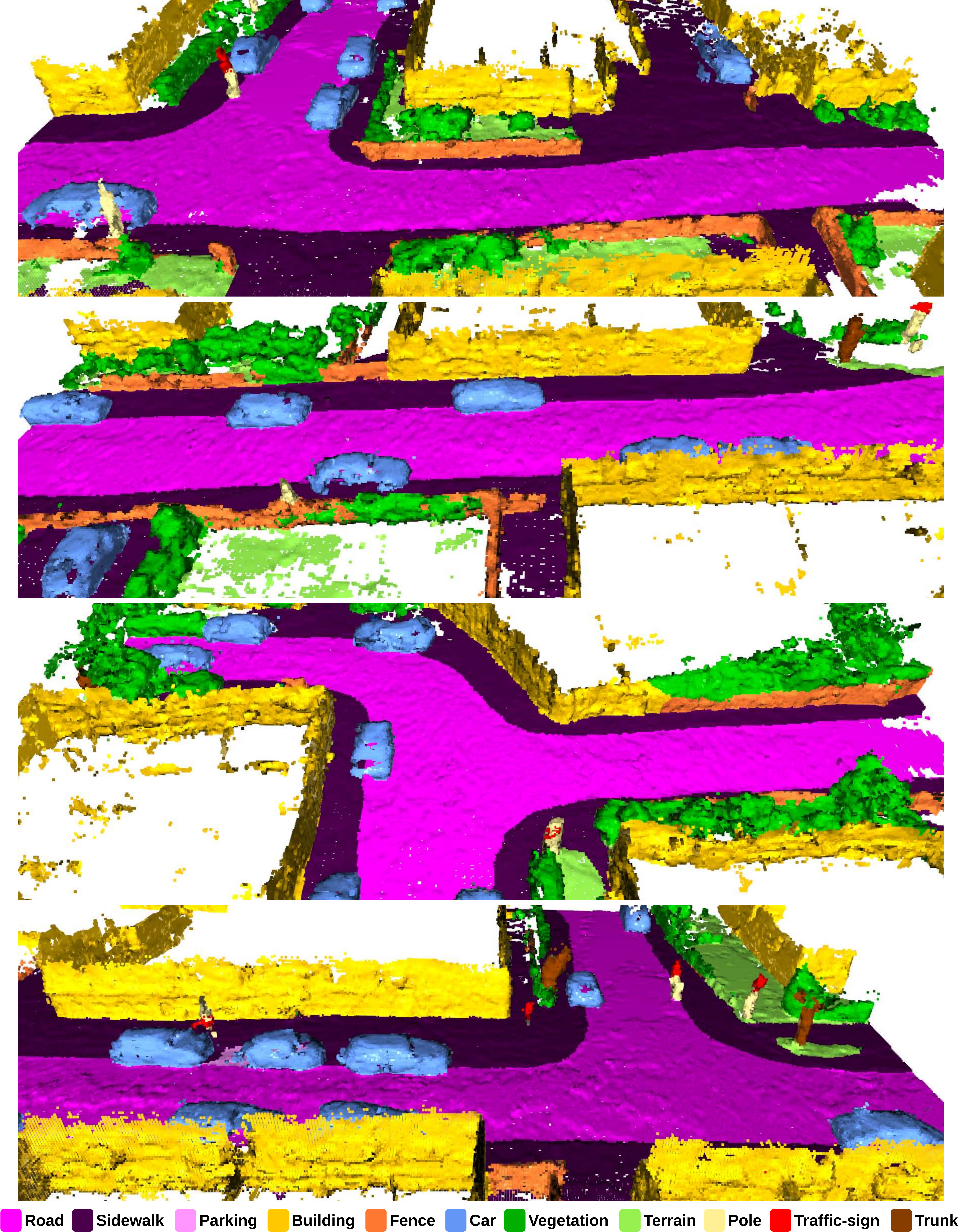}
    \caption{Unconditional scenes generated by our method.}
    \label{fig:qualitative2}
\end{figure*}

\begin{figure*}[t]
    \centering
    \includegraphics[width=0.99\linewidth]{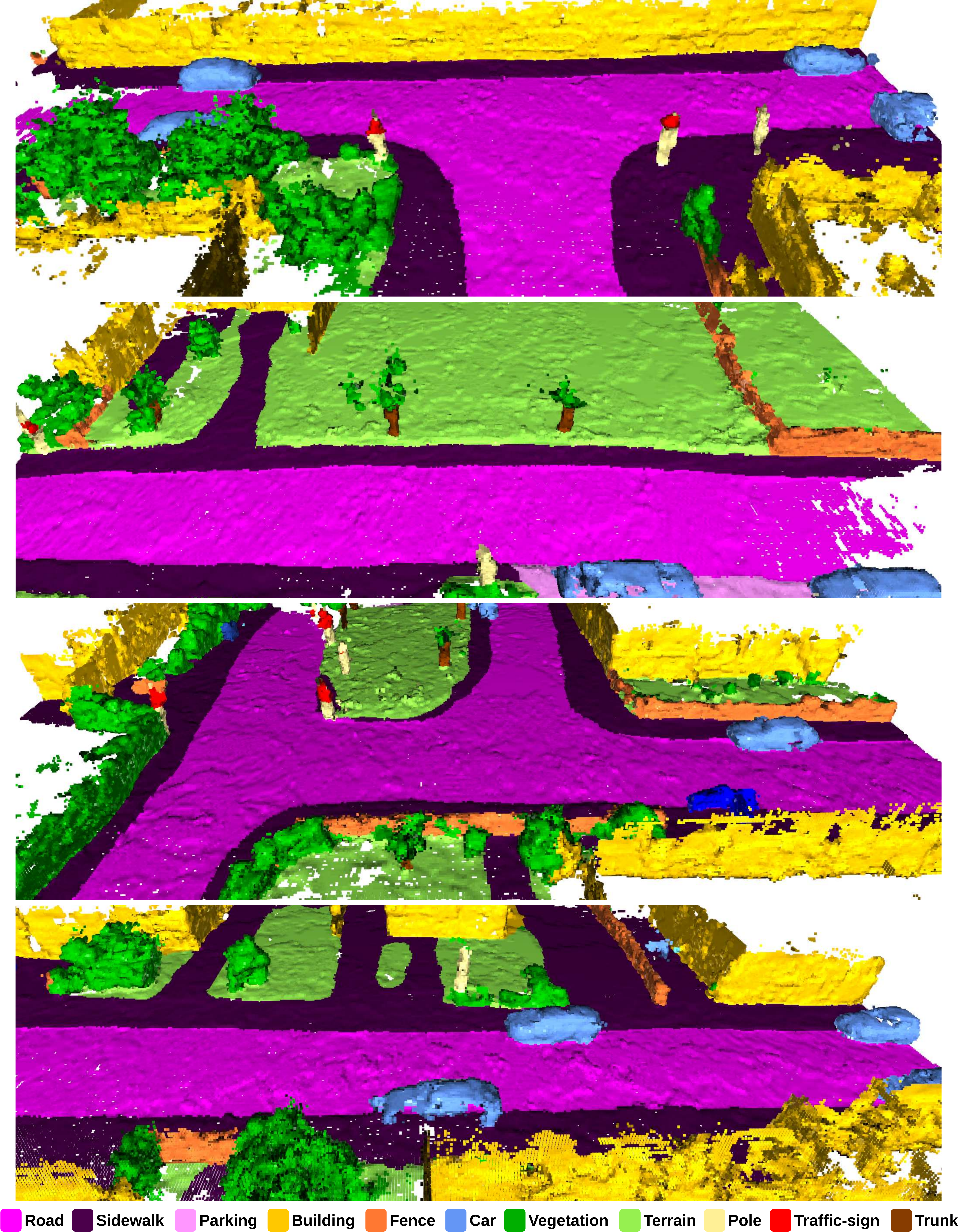}
    \caption{Unconditional scenes generated by our method.}
    \label{fig:qualitative3}
\end{figure*}

\begin{figure*}[t]
    \centering
    \includegraphics[width=0.99\linewidth]{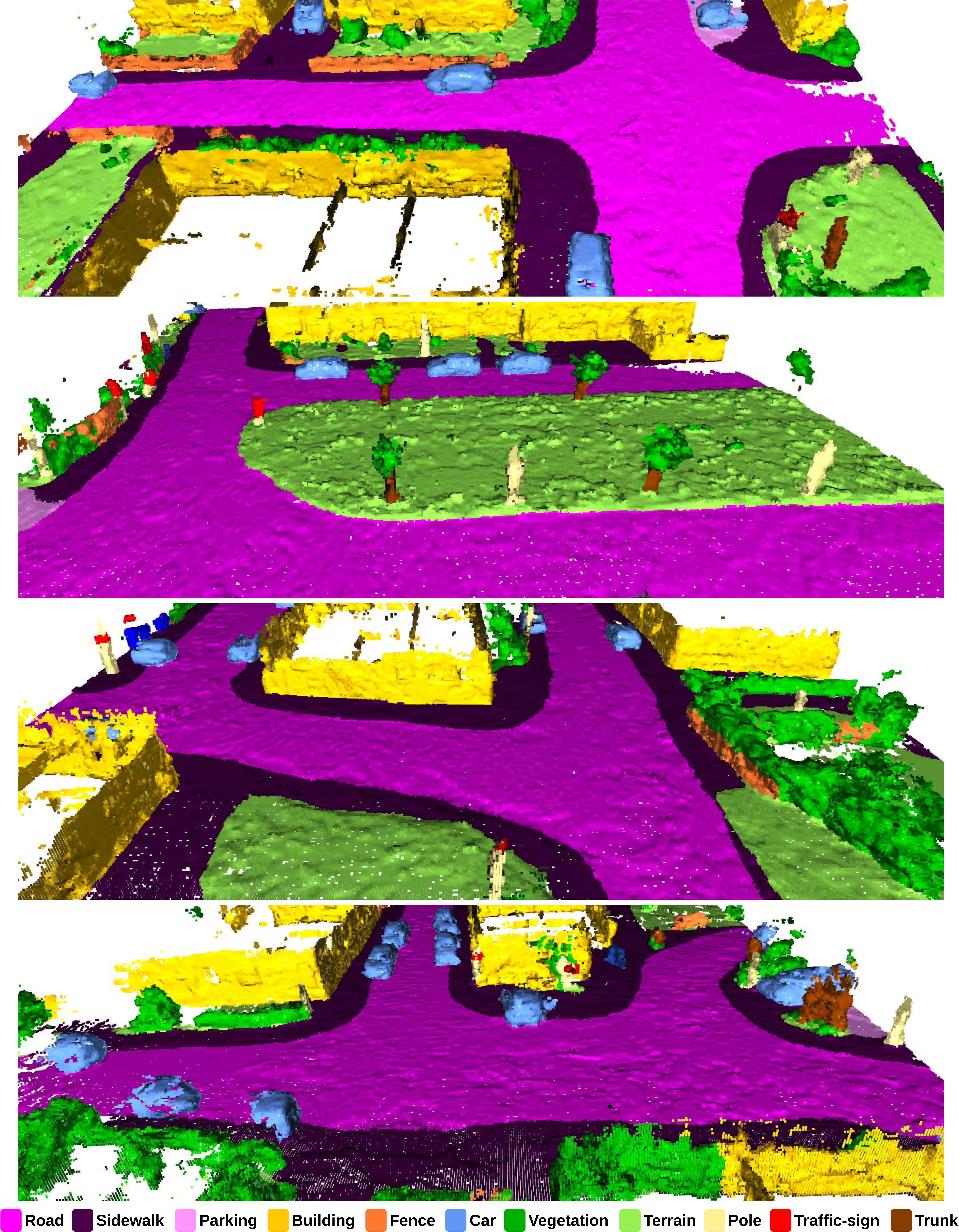}
    \caption{Unconditional scenes generated by our method.}
    \label{fig:qualitative4}
\end{figure*}

\begin{figure*}[t]
    \centering
    \includegraphics[width=0.99\linewidth]{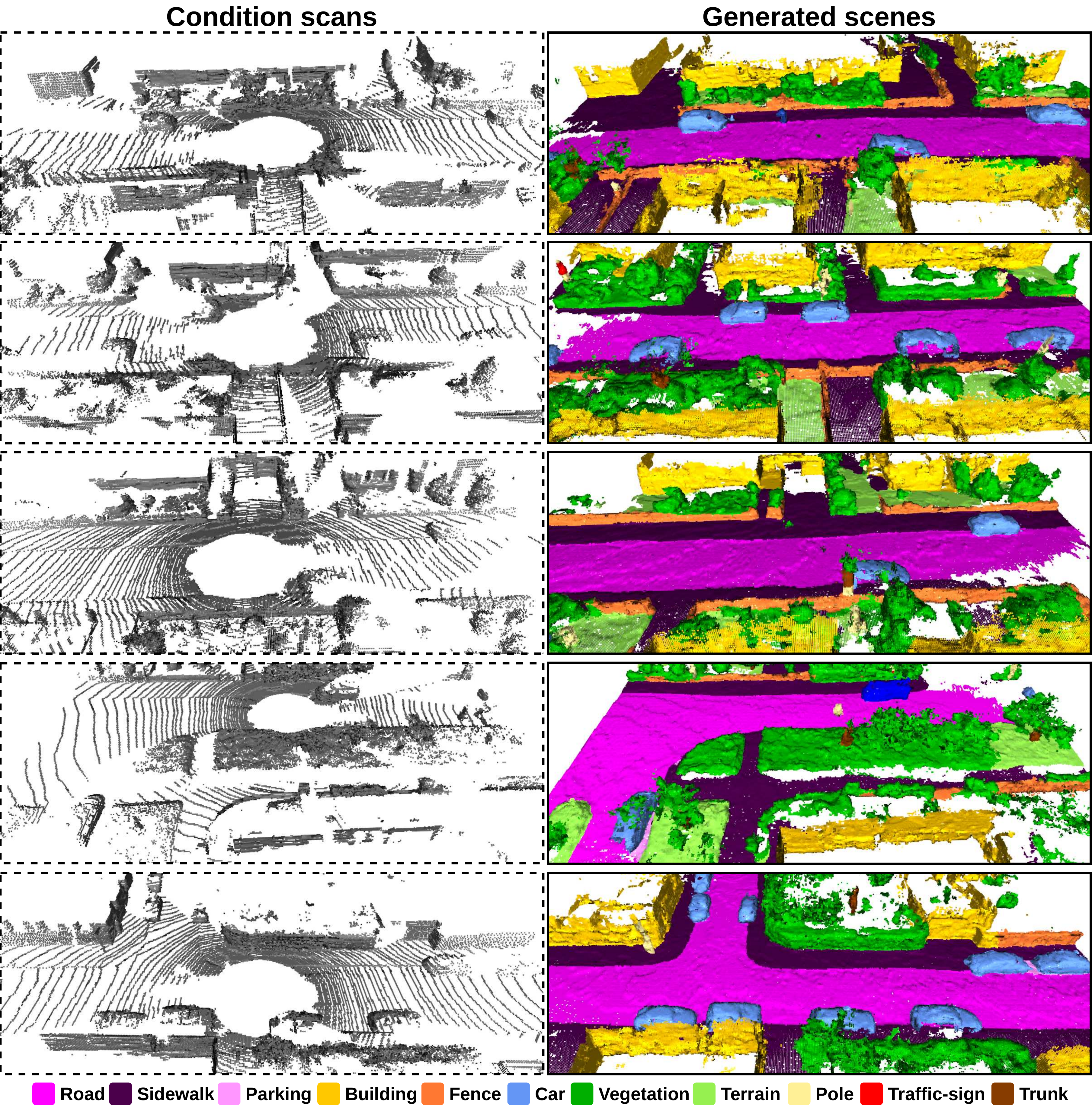}
    \caption{Generated scenes conditioned with unseen scans from KITTI-360 dataset~\cite{liao2022pami}.}
    \label{fig:qualitative_360}
\end{figure*}

\begin{figure*}[t]
    \centering
    \includegraphics[width=0.99\linewidth]{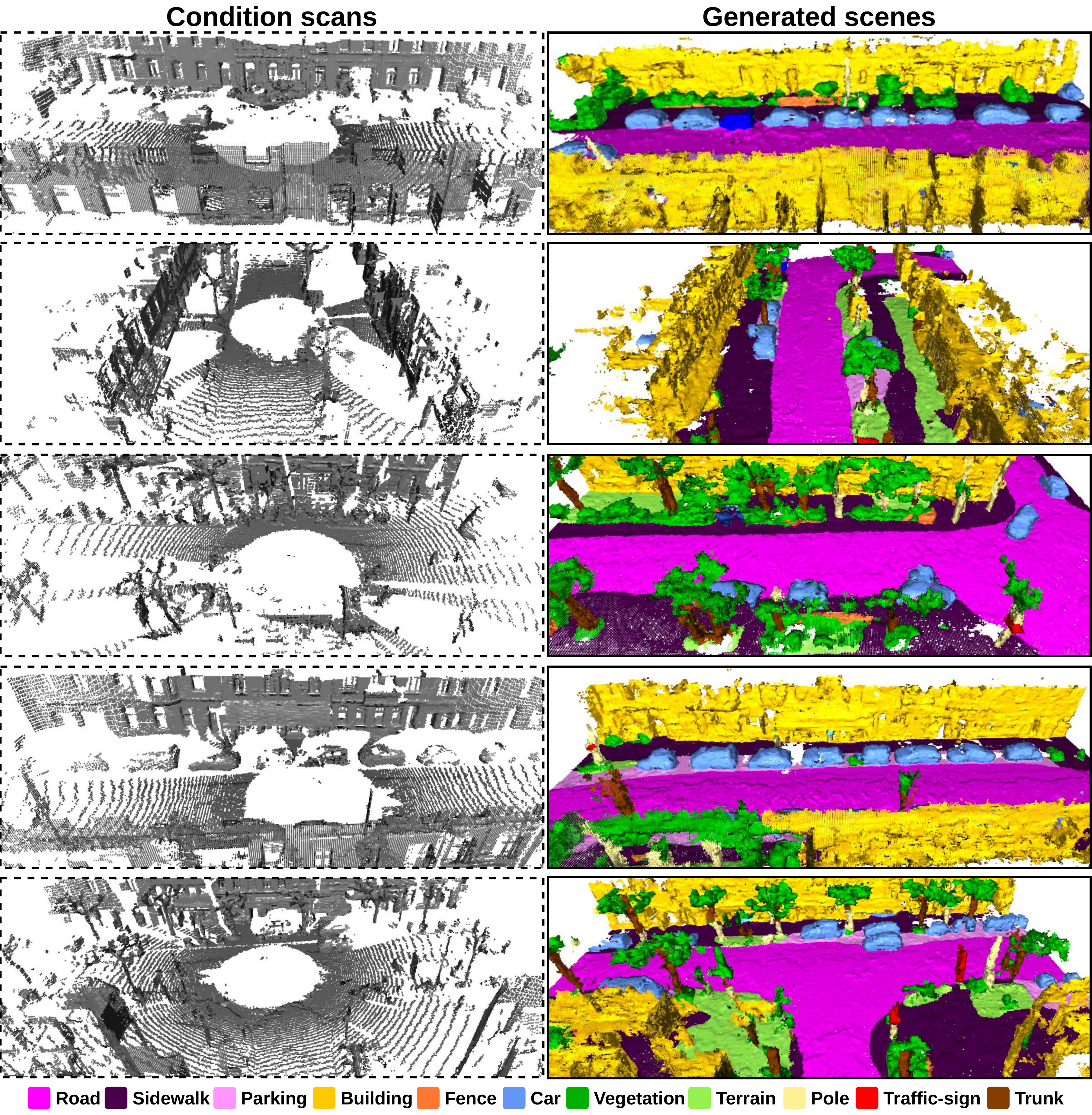}
    \caption{Generated scenes conditioned with unseen scans from our data collected with an Ouster OS-1 $128$ beam LiDAR.}
    \label{fig:qualitative_ipbcar}
\end{figure*}

In this section, we show further examples of scenes generated by our method. \cref{fig:qualitative2,fig:qualitative3,fig:qualitative4} shows the unconditionally generated scene examples. As shown, the
randomly generated point clouds present different scenarios. This variability agrees with the discussion in Sec. 4.2 where the training with a mixed
set of samples from real and synthetic scenes performed better than using only real scenes. The synthetic scenes add variability to the data
compared to the real training set since the real scenes were collected sequentially, where consecutive scenes may have few changes. In contrast, by using
randomly generated scenes, each individual sample will be different from each other, adding variability to the data, thus improving the performance of the
model when trained with this synthetic training set. Regarding the conditioned scene generation, \cref{fig:qualitative_360,fig:qualitative_ipbcar} show examples of scenes generated conditioned to LiDAR scans from KiTTI-360~\cite{liao2022pami} and to our own collected data, respectively.
In those examples, we can notice that the model is able to generate reasonable scenes according to the scan used as condition. Such conditioning also works for point clouds collected with a LiDAR sensor different from the one used
to train the model, as seen in \cref{fig:qualitative_ipbcar} from the scenes generated conditioned to Ouster LiDAR collected by us.

\section{Limitations}
\label{sec:limitations}

\begin{figure*}[h]
  \centering\includegraphics[width=0.95\linewidth]{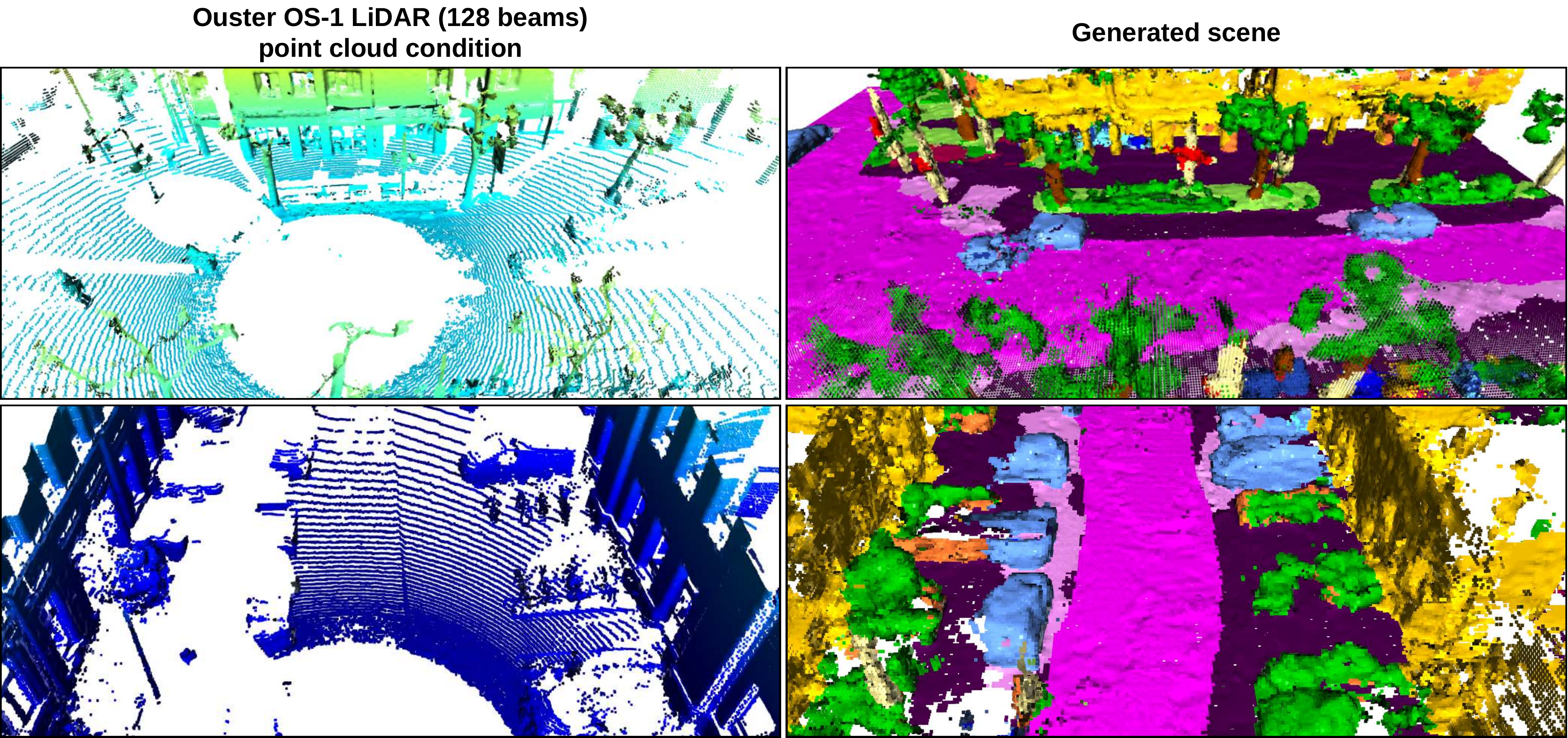}\vspace{-0.1cm}
  \caption{Failure case example. Condition point cloud collected with a LiDAR (Ouster OS-1 128 beams) with a different resolution than the one used to collect the training data (Velodyne HDL-64E 64 beams).}
  \label{fig:failure_case}
\end{figure*}

\reviewed{In this section, we discuss the limitations of our proposed method. \cref{fig:failure_case} presents examples of failure cases of the conditional generation samples from our method. Our conditional model was trained on SemanticKITTI scans, collected with a Velodyne HDL-64E with 64 beams. Our data was collected with an Ouster OS-1 with 128 beams. Both LiDARs have different numbers of beams and laser beam patterns, leading to a domain gap between the scans used during training and those used during conditioning. As shown in \cref{fig:failure_case}, when generating scenes from point clouds collected with different LiDARs, artifacts can be generated, leading to non-realistic semantic scenes. Still, this limitation can be tackled with a curation step, as discussed in the main paper.}

{\small
\bibliographystyle{IEEEtranS}
\bibliography{glorified, new}
}